%% file: mesa.tex
\documentclass{article}

\pdfoutput=1

    \usepackage[nonatbib,preprint]{neurips_2020}

\usepackage[utf8]{inputenc} % allow utf-8 input
\usepackage[T1]{fontenc}    % use 8-bit T1 fonts
\usepackage{hyperref}       % hyperlinks
\usepackage{url}            % simple URL typesetting
\usepackage{booktabs}       % professional-quality tables
\usepackage{amsfonts}       % blackboard math symbols
\usepackage{nicefrac}       % compact symbols for 1/2, etc.
\usepackage{microtype}      % microtypography

\usepackage{threeparttable}
\usepackage{multicol}
\usepackage[normalem]{ulem}
\usepackage{amsmath}
\usepackage{amssymb}
\usepackage{amsfonts}
\usepackage[utf8]{inputenc}
\usepackage{geometry}
\usepackage{mathtools}
\usepackage{comment}
\usepackage{graphicx}
\usepackage{pifont}
\usepackage{multirow}
\usepackage{color}
\usepackage{diagbox}
\usepackage{subfigure}
\usepackage{fitbox}
\usepackage{fancybox}
\usepackage{xcolor}
\usepackage{caption}
\usepackage{algorithm}
\usepackage[noend]{algpseudocode}
\usepackage{wrapfig}

\title{MESA: Boost Ensemble Imbalanced Learning with MEta-SAmpler}

\author{%
  Zhining Liu \\ % \thanks{} \\
  Jilin University\\
  \texttt{znliu19@mails.jlu.edu.cn} \\
  \And
  Pengfei Wei \\
  National University of Singapore \\
  \texttt{dcsweip@nus.edu.sg} \\
  \And
  Jing Jiang \\
  University of Technology Sydney \\
  \texttt{jing.jiang@uts.edu.au} \\
  \And
  Wei Cao \\
  Microsoft Research \\
  \texttt{weicao@microsoft.com} \\
  \And
  Jiang Bian \\
  Microsoft Research \\
  \texttt{jiang.bian@microsoft.com} \\
  \And
  Yi Chang \\
  Jilin University\\
  \texttt{yichang@jlu.edu.cn} \\
}

\begin{document}

\maketitle

\newcommand{\codeurl}{{\tt https://github.com/ZhiningLiu1998/mesa}}

\begin{abstract}
  {\it Imbalanced learning} (IL), i.e., learning unbiased models from class-imbalanced data, is a challenging problem.
  Typical IL methods including resampling and reweighting were designed based on some heuristic assumptions. 
  They often suffer from unstable performance, poor applicability, and high computational cost in complex tasks where their assumptions do not hold.
  % Some recent research efforts apply meta-learning to pursue better IL but are confined to be co-optimized with deep neural networks (DNNs).
  % This greatly limited their application to other learning algorithms (e.g., tree-based models) which are better choices in traditional tasks (e.g., small/unstructured/tabular data).
  In this paper, we introduce a novel ensemble IL framework named {\sc Mesa}.
  It adaptively resamples the training set in iterations to get multiple classifiers and forms a cascade ensemble model. 
  {\sc Mesa} directly learns the sampling strategy from data to optimize the final metric beyond following random heuristics.
  Moreover, unlike prevailing meta-learning-based IL solutions, we decouple the model-training and meta-training in {\sc Mesa} by independently train the meta-sampler over task-agnostic meta-data.
  This makes {\sc Mesa} generally applicable to most of the existing learning models and the meta-sampler can be efficiently applied to new tasks.
  Extensive experiments on both synthetic and real-world tasks demonstrate the effectiveness, robustness, and transferability of {\sc Mesa}.
  Our code is available at \codeurl.
\end{abstract}

\input{sections/introduction}
\input{sections/related-work}
\input{sections/meta-sampler}
\input{sections/experiments}
\input{sections/conclusion}
\input{sections/broader-impact}

\begin{ack}
  We thank anonymous reviewers for their constructive suggestions on improving the paper.
\end{ack}

\bibliographystyle{plain}
\bibliography{mesa-ref}

\newpage
\appendix
\input{sections/reproducibility}

\end{document}

%% file: sections/introduction.tex
\section{Introduction}

{\it Class imbalance}, due to the naturally-skewed class distributions, has been widely observed in many real-world applications such as click prediction, fraud detection, and medical diagnosis~\cite{graepel2010ctr,haixiang2017learning-from-imb-review,japkowicz2002systematic-study}.
Canonical classification algorithms usually induce the bias, i.e., perform well in terms of global accuracy but poorly on the minority class, in solving class imbalance problems.
However, the minority class commonly yields higher interests from both learning and practical perspectives~\cite{he2008overview,he2013overview}.

Typical imbalanced learning (IL) algorithms attempt to eliminate the bias through data {\it resampling} \cite{chawla2002smote,han2005borderline-smote,he2008adasyn,laurikkala2001ncr,mani2003nearmiss} or {\it reweighting} \cite{lin2017focalloss,liu2006cost-sensitive-imbalance,shrivastava2016hard-example-mining} in the learning process.
More recently, ensemble learning is incorporated to reduce the variance introduced by resampling or reweighting and has achieved satisfactory performance~\cite{krawczyk2016learning}.
In practice, however, all these methods have been observed to suffer from three major limitations: 
(I) unstable performance due to the sensitivity to outliers, (II) poor applicability because of the prerequisite of domain experts to hand-craft the cost matrix, and (III) high cost of computing the distance between instances.

Regardless the computational issue, we attribute the unsatisfactory performance of traditional IL methods to the validity of heuristic assumptions made on training data.
For instance, some methods~\cite{chawla2003smoteboost,freund1997adaboost,liu2009ee-bc,seiffert2010rusboost} assume instances with higher training errors are more informative for learning.
However, misclassification may be caused by outliers, and error reinforcement arises in this case with the above assumption. 
Another widely used assumption is that generating synthetic samples around minority instances helps with learning~\cite{chawla2003smoteboost,chen2010ramoboost,wang2009smotebagging}.
This assumption only holds when the minority data is well clustered and sufficiently discriminative. 
If the training data is extremely imbalanced or with many corrupted labels, the minority class would be poorly represented and lack a clear structure. 
In this case, working under this assumption severely jeopardizes the performance. 

Henceforth, it is much more desired to develop an adaptive IL framework that is capable of handling complex real-world tasks without intuitive assumptions.
Inspired by the recent developments in meta-learning~\cite{lake2015meta-learning}, we propose to achieve the meta-learning mechanism in ensemble imbalanced learning (EIL) framework.
In fact, some preliminary efforts~\cite{peng2019trainable-under-sampling,ren2018learning-to-reweight,han2018meta-weight-net} have investigated the potential of applying meta-learning to IL problems.
Nonetheless, these works have limited capability of generalization because of the model-dependent optimization process.
Their meta-learners are confined to be co-optimized with a single DNN, which greatly limits their application to other learning models (e.g., tree-based models) as well as deployment into the more powerful EIL framework.

In this paper, we propose a generic EIL framework {\sc Mesa} that automatically learns its strategy, i.e., the meta-sampler, from data towards optimizing imbalanced classification.
The main idea is to model a meta-sampler that serves as an adaptive under-sampling solution embedded in the iterative ensemble training process.
In each iteration, it takes the current state of ensemble training (i.e., the classification error distribution on both the training and validation sets) as its input.
Based on this, the meta-sampler selects a subset to train a new base classifier and then adds it to the ensemble, a new state can thus be obtained.
We expect the meta-sampler to maximize the final generalization performance by learning from such interactions.
To this end, we use reinforcement learning (RL) to solve the non-differentiable optimization problem of the meta-sampler.
To summarize, this paper makes the following contributions.
(I) We propose {\sc Mesa}, a generic EIL framework that demonstrates superior performance by automatically learning an adaptive under-sampling strategy from data.
(II) We carry out a preliminary exploration of extracting and using cross-task meta-information in EIL systems.
The usage of such meta-information gives the meta-sampler cross-task transferability.
A pretrained meta-sampler can be directly applied to new tasks, thereby greatly reducing the computational cost brought about by meta-training.
(III) Unlike prevailing methods whose meta-learners were designed to be co-optimized with a specific learning model (i.e, DNN) during training, we decoupled the model-training and meta-training process in {\sc Mesa}.
This makes our framework generally applicable to most of the statistical and non-statistical learning models (e.g., decision tree, Na\"ive Bayes, k-nearest neighbor classifier).

%% file: sections/related-work.tex
\section{Related Work}
\label{section:background}

Fernández et al.~\cite{albert02018experiment}, Guo et al.~\cite{haixiang2017learning-from-imb-review}, and He et al.~\cite{he2008overview,he2013overview} provided systematic reviews of algorithms and applications of imbalanced learning.
In this paper, we focus on \emph{binary imbalanced classification} problem, which is one of the most widely studied problem setting~\cite{haixiang2017learning-from-imb-review,krawczyk2016learning} in imbalanced learning.
Such a problem extensively exists in practical applications, e.g., fraud detection (fraud vs. normal), medical diagnosis (sick vs. healthy), and cybersecurity (intrusion vs. user connection).
We mainly review existing works on this problem as follows.

{\bf Resampling} Resampling methods focus on modifying the training set to balance the class distribution (i.e., over/under-sampling~\cite{chawla2002smote,han2005borderline-smote,he2008adasyn,mani2003nearmiss,smith2014instance-complexity}) or filter noise (i.e., cleaning resampling~\cite{laurikkala2001ncr,tomek1976tomeklink}).
Random resampling usually leads to severe information loss or overfishing, hence many advanced methods explore distance information to guide their sampling process~\cite{haixiang2017learning-from-imb-review}.
However, calculating the distance between instances is computationally expensive on large-scale datasets, and such strategies may even fail to work when the data does not fit their assumptions.

{\bf Reweighting} Reweighting methods assign different weights to different instances to alleviate a classifier's bias towards majority groups (e.g.,~\cite{chai2004csnb,freund1997adaboost,ling2004csdt,liu2006cost-sensitive-imbalance}).
Many recent reweighting methods such as FocalLoss~\cite{lin2017focalloss} and GHM~\cite{li2019gradient-harmonize} are specifically designed for DNN loss function engineering.
Class-level reweighting such as cost-sensitive learning~\cite{liu2006cost-sensitive-imbalance} is more versatile but requires a cost matrix given by domain experts beforehand, which is usually infeasible in practice.

{\bf Ensemble Methods.}
Ensemble imbalanced learning (EIL) is known to effectively improve typical IL solutions by combining the outputs of multiple classifiers (e.g.,~\cite{chawla2003smoteboost,liu2009ee-bc,liu2019self-paced-ensemble,seiffert2010rusboost,wang2009smotebagging}).
These EIL approaches prove to be highly competitive~\cite{krawczyk2016learning} and thus gain increasing popularity~\cite{haixiang2017learning-from-imb-review} in IL.
However, most of the them are straight combinations of a resampling/reweighting solution and an ensemble learning framework, e.g., {\sc Smote~\cite{chawla2002smote}+AdaBoost~\cite{freund1997adaboost}=SmoteBoost~\cite{chawla2003smoteboost}}.
Consequently, although EIL techniques effectively lower the variance introduced by resampling/reweighting, these methods still suffer from unsatisfactory performance due to their heuristic-based designs.

{\bf Meta-learning Methods.}
Inspired by recent meta-learning developments~\cite{finn2017model-agnostic-meta-learning,lake2015meta-learning}, there are some studies that adapt meta-learning to solve IL problem.
Typical methods include Learning to Teach~\cite{wu2018learning2teach} that learns a dynamic loss function, MentorNet~\cite{jiang2017mentornet} that learns a mini-batch curriculum, and L2RW~\cite{ren2018learning-to-reweight}/Meta-Weight-Net~\cite{han2018meta-weight-net} that learn an implicit/explicit data weighting function.
Nonetheless, all these methods are confined to be co-optimized with a DNN by gradient descent.
As the success of deep learning relies on the massive training data, mainly from the well-structured data domain like computer vision and natural language processing, the applications of these methods to other learning models (e.g., tree-based models and their ensemble variants like gradient boosting machine) in traditional classification tasks (e.g., small/unstructured/tabular data) are highly constrained. 

We present a comprehensive comparison of existing IL solutions for binary imbalanced classification problem with our {\sc Mesa} in Table~\ref{table:comparison}.
Compared with other methods, {\sc Mesa} aims to learn a resampling strategy directly from data.
It is able to perform quick and adaptive resampling as no distance computing, domain knowledge, or related heuristics are involved in the resampling process.

\newcommand{\yes}{\ding{51}}
\newcommand{\no}{\ding{55}}
\newcommand{\yesno}{\textcolor{black}{\ding{51}}{\textcolor{black}{\kern-0.65em\ding{55}}}}
\begin{table*}[t]
\centering
\tiny
\caption{
    Comparisons of {\sc Mesa} with existing imbalanced learning methods, note that $\mathcal{|N|} \gg \mathcal{|P|}$.
    }
\label{table:comparison}
\begin{threeparttable}
\resizebox{\linewidth}{!}{
\begin{tabular}{c|c|ccccc}
\toprule
\multirow{2}*{Category\tnote{*}} & \multirow{2}*{Representative(s)} &
    Sample      & Distance-based   & Domain kno-  & Robust to noi-& \multirow{2}*{Requirements}\\
& & efficiency  & resampling cost  & wledge free? & ses/outliers? & \\
\midrule
RW & \cite{ling2004csdt}, \cite{chai2004csnb} & $\mathcal{O(|P|+|N|)}$ &
\no & \no & \yesno & cost matrix set by domain experts\\

US  & \cite{mani2003nearmiss}, \cite{smith2014instance-complexity} & $\mathcal{O}(2|\mathcal{P}|)$ &
$\mathcal{O}(|\mathcal{P}|)$ & \yes & \no & well-defined distance metric \\

OS  & \cite{chawla2002smote}, \cite{he2008adasyn} & $\mathcal{O}(2|\mathcal{N}|)$ &
$\mathcal{O}(|\mathcal{P}|)$ & \yes & \no & well-defined distance metric \\

CS  & \cite{wilson1972enn}, \cite{tomek1976allknn}& $\mathcal{O(|P|+|N|)}$ &
$\mathcal{O}(|\mathcal{P}|\cdot|\mathcal{N}|)$ & \yes & \yes & well-defined distance metric \\

OS+CS  & \cite{batista2004smoteenn}, \cite{batista2003smotetomek} & $\mathcal{O}(2|\mathcal{N}|)$ &
$\mathcal{O}(|\mathcal{P}|\cdot|\mathcal{N}|)$ & \yes & \yes & well-defined distance metric \\

\midrule
IE+RW & \cite{freund1997adaboost}, \cite{sun2007cost-boost} & $\mathcal{O}(k(\mathcal{|P|+|N|}))$ &
\no & \no & \no & cost matrix set by domain experts\\

PE+US  & \cite{barandela2003underbagging}, \cite{liu2009ee-bc} & $\mathcal{O}(2k|\mathcal{P}|)$ &
\no & \yes & \yes & - \\

PE+OS  & \cite{wang2009smotebagging} & $\mathcal{O}(2k|\mathcal{N}|)$ &
$\mathcal{O}(2k|\mathcal{P}|)$ & \yes & \yes & well-defined distance metric \\

IE+RW+US  & \cite{seiffert2010rusboost} & $\mathcal{O}(2k|\mathcal{P}|)$ &
\no & \yes & \no & - \\

IE+RW+OS  & \cite{chawla2003smoteboost} & $\mathcal{O}(2k|\mathcal{N}|)$ &
$\mathcal{O}(2k|\mathcal{P}|)$ & \yes & \no & well-defined distance metric\\

\midrule
ML  & \cite{han2018meta-weight-net}, \cite{ren2018learning-to-reweight}, \cite{wu2018learning2teach} & $\mathcal{O(|P|+|N|)}$ &
\no & \yesno & \yes & co-optimized with DNN only \\

IE+ML  & MESA(ours) & $\mathcal{O}(2k|\mathcal{P}|)$ &
\no & \yes & \yes & independent meta-training \\

\bottomrule
\end{tabular}}

\begin{tablenotes}
    \tiny
    \item[*] reweighting (RW), under-sampling (US), over-sampling (OS), cleaning-sampling (CS), iterative ensemble (IE), parallel ensemble (PE), meta-learning (ML).
\end{tablenotes}
\end{threeparttable}
\end{table*}

%% file: sections/meta-sampler.tex
\section{The proposed \scshape{Mesa} framework}
\label{section:meta-sampler}

\begin{figure}[t]
  \centering
  \includegraphics[width=1\linewidth]{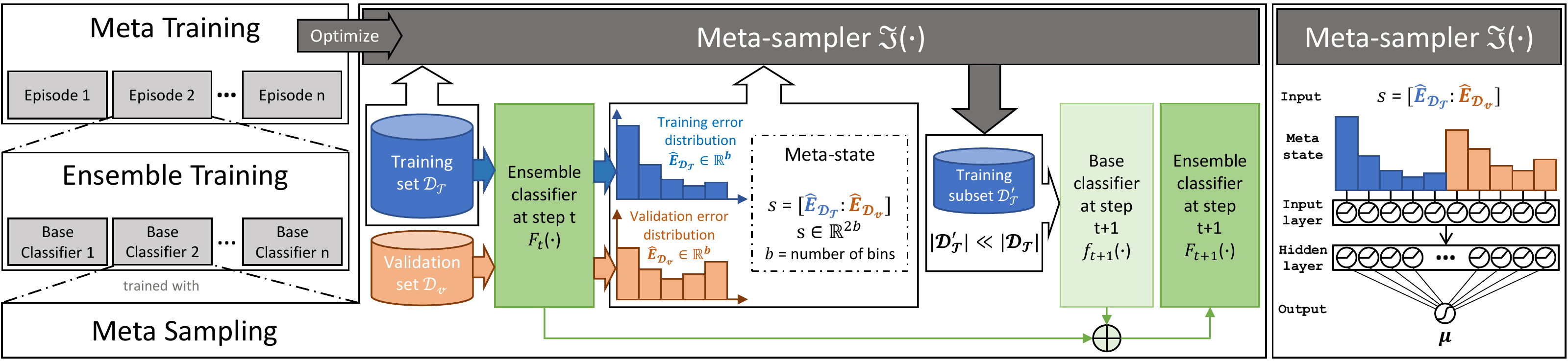}
  \caption{Overview of the proposed {\sc Mesa} Framework. Best viewed in color.}
  \label{fig:overview}
\end{figure}

In order to take advantage of both ensemble learning and meta-learning, we propose a novel EIL framework named {\sc Mesa} that works with a meta-sampler. 
As shown in Fig.~\ref{fig:overview}, {\sc Mesa} consists of three parts: {\it meta-sampling} as well as {\it ensemble training} to build ensemble classifiers, and {\it meta-training} to optimize the meta-sampler. 
We will describe them respectively in this section. 

Specifically, {\sc Mesa} is designed to: 
(I) perform resampling based on meta-information to further boost the performance of ensemble classifiers; 
(II) decouple model-training and meta-training for general applicability to different classifiers;
(III) train the meta-sampler over task-agnostic meta-data for cross-task transferability and reducing meta-training cost on new tasks.

{\bf Notations.}
Let $\mathcal{X}:\mathbb{R}^d$ be the input feature space and $\mathcal{Y}:\{0,1\}$ be the label space. 
An instance is represented by $(x,y)$, where $x \in \mathcal{X}$, $y \in \mathcal{Y}$. 
Without loss of generality, we always assume that the minority class is positive. 
Given an imbalanced dataset $\mathcal{D}: \{(x_1,y_1), (x_2,y_2), \cdots, (x_n,y_n)\}$, the minority set is $\mathcal{P}: \{(x, y)\ |\ y = 1, (x, y) \in \mathcal{D}\}$ and the majority set is $\mathcal{N}: \{(x, y)\ |\ y = 0, (x, y) \in \mathcal{D}\}$. 
For highly imbalanced data we have $|\mathcal{N}| \gg |\mathcal{P}|$. 
We use $f: x \to [0, 1]$ to denote a single classifier and $F_k: x \to [0, 1]$ to denote an ensemble classifier that is formed by $k$ base classifiers. 
We use $\mathcal{D}_{\tau}$ and $\mathcal{D}_{v}$ to represent the training set and validation set, respectively.

{\bf Meta-state.}
As mentioned before, we expect to find a task-agnostic representation that can provide the meta-sampler with the information of the ensemble training process.
Motivated by the concept of ``gradient/hardness distribution'' from~\cite{li2019gradient-harmonize,liu2019self-paced-ensemble}, we introduce the histogram distribution of the training and validation errors as the meta-state of the ensemble training system.

Formally, given an data instance $(x,y)$ and an ensemble classifier $F_{t}(\cdot)$, the classification error $e$ is defined as the absolute difference between the predicted probability of $x$ being positive and the ground truth label $y$, i.e., $|F_{t}(x)-y|$.
Suppose the error distribution on dataset $\mathcal{D}$ is $E_{\mathcal{D}}$, then the error distribution approximated by histogram is given by a vector $\widehat{E}_{\mathcal{D}} \in \mathbb{R}^b$, where $b$ is the number of bins in the histogram. 
Specifically, the $i$-th component of vector $\widehat{E}_{\mathcal{D}}$  can be computed as follows\footnote{To avoid confusion, in Eq.~\ref{eq:error-distribution}, we use $|\cdot|$ and $abs(\cdot)$ to denote cardinality and absolute value, respectively.}: 
\begin{equation}
    \label{eq:error-distribution}
    \widehat{E}_{\mathcal{D}}^{i} = \frac{|\{(x,y)\ |\ \frac{i-1}{b} \le abs(F_{t}(x)-y) < \frac{i}{b}\ ,(x,y) \in \mathcal{D} \}|}{|\mathcal{D}|}, 1 \le i \le b. % i \in \{1, 2, \dots, b\}
\end{equation}
After concatenating the error distribution vectors on training and validation set, we have the meta-state: 
\begin{equation}
    \label{eq:meta-state}
    s = [\widehat{E}_{\mathcal{D}_{\tau}}:\widehat{E}_{\mathcal{D}_{v}}] \in \mathbb{R}^{2b}.
\end{equation}

\begin{figure}[t]
  \begin{minipage}{0.48\linewidth}
    \centering
    \includegraphics[width=1\linewidth]{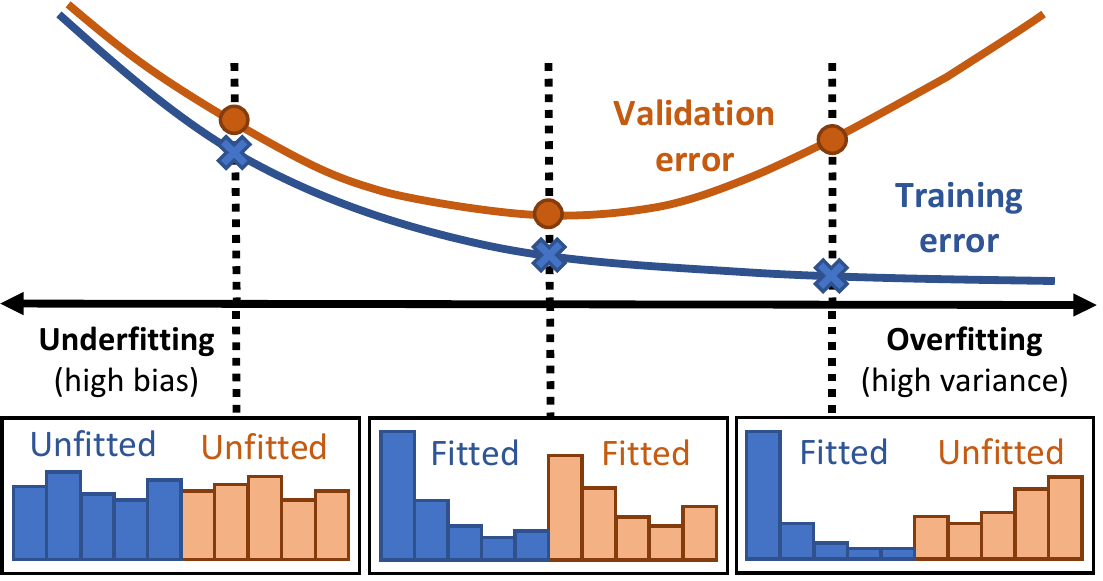}
  \end{minipage}
  \hspace{\fill}
  \begin{minipage}{0.51\linewidth}
    \vspace{5pt}
    \caption{
      Some examples of different meta-states ($s = [\widehat{E}_{\mathcal{D}_{\tau}}:\widehat{E}_{\mathcal{D}_{v}}]$) and their corresponding ensemble training states.
      The meta-state reflects how well the current classifier fits on the training set, and how well it generalizes to unseen validation data.
      Note that such representation is independent of properties of the specific task (e.g., dataset size, feature space) thus can be used to support the meta-sampler to perform adaptive resampling across different tasks.
    }
    \label{fig:meta-state}
  \end{minipage}
\end{figure}

Intuitively, the histogram error distribution $\widehat{E}_{\mathcal{D}}$ shows how well the given classifier fits the dataset $\mathcal{D}$.
When $b=2$, it reports the accuracy score in $\widehat{E}_{\mathcal{D}}^1$ and misclassification rate in $\widehat{E}_{\mathcal{D}}^2$ (classification threshold is 0.5).
With $b>2$, it shows the distribution of ``easy'' samples (with errors close to 0) and ``hard'' samples (with errors close to 1) in finer granularity, thus contains more information to guide the resampling process.
Moreover, since we consider both the training and validation set, the meta-state also provides the meta-sampler with information about bias/variance of the current ensemble model and thus supporting its decision. 
We show some illustrative examples in Fig.~\ref{fig:meta-state}.

{\bf Meta Sampling.}
Making instance-level decisions by using a complex meta-sampler (e.g., set a large output layer or use recurrent neural network) is extremely time-consuming as the complexity of a single update $C_{u}$ is $\mathcal{O}(|\mathcal{D}|)$.
Besides, complex model architecture also brings extra memory cost and hardship in optimization.
To make {\sc Mesa} more concise and efficient, we use a Gaussian function trick to simplify the meta-sampling process and the sampler itself, reducing $C_{u}$ from $\mathcal{O}(|\mathcal{D}|)$ to $\mathcal{O}(1)$.

Specifically, let $\Im$ denote the meta-sampler, it outputs a scalar $\mu \in [0,1]$ based on the input meta-state $s$, i.e., $\mu \thicksim \Im(\mu|s)$.
We then apply a Gaussian function $g_{\mu,\sigma}(x)$ over each instance's classification error to decide its (unnormalized) sampling weight, where $g_{\mu,\sigma}(x)$ is defined as:
\begin{equation}
    \label{eq:gaussian-function}
    g_{\mu,\sigma}(x) = \frac{1}{\sigma \sqrt{2 \pi}}e^{-\frac{1}{2}(\frac{x-\mu}{\sigma})^2}.
\end{equation}
Note that in Eq.~\ref{eq:gaussian-function}, $e$ is the Euler's number, $\mu \in [0,1]$ is given by the meta-sampler and $\sigma$ is a hyper-parameter. 
Please refer to Section~\ref{section:hyper-parameter} for discussions and guidelines about our hyper-parameter setting. 
The above meta-sampling procedure $\texttt{Sample}(~\cdot~; F, \mu, \sigma)$ is summarized in Algorithm~\ref{alg:meta-sampling}.

\begin{table}
  \begin{minipage}{0.48\linewidth}
    \begin{algorithm}[H]
      \caption{$\text{\tt Sample}(\mathcal{D}_{\tau}; F, \mu, \sigma)$}
      \label{alg:meta-sampling}
      \begin{algorithmic}[1]
        \Require $\mathcal{D}_{\tau}$, $F$, $\mu$, $\sigma$
        \State Initialization: derive majority set $\mathcal{P}_{\tau}$ and minority set $\mathcal{N}_{\tau}$ from $\mathcal{D}_{\tau}$\;
        \State Assign each $(x_i,y_i)$ in $\mathcal{N}_{\tau}$ with weight: $$w_i = \frac{g_{\mu,\sigma}(|F(x_i)-y_i|)}{\sum_{(x_j, y_j) \in \mathcal{N}_{\tau}} g_{\mu,\sigma}(|F(x_j)-y_j|)}$$
        \State Sample majority subset $\mathcal{N}_{\tau}^{'}$ from $\mathcal{N}_{\tau}$ w.r.t. sampling weights $w$, where $|\mathcal{N}_{\tau}^{'}| = |\mathcal{P}_{\tau}|$\;
        \State \Return balanced subset $\mathcal{D}_{\tau}^{'} = \mathcal{N}_{\tau}^{'} \cup \mathcal{P}_{\tau}$
      \end{algorithmic}
    \end{algorithm}
  \end{minipage}
  \hspace{10pt}
  \begin{minipage}{0.48\linewidth}
    \begin{algorithm}[H]
      \caption{Ensemble training in {\sc Mesa}}
      \label{alg:ensemble-training}
      \begin{algorithmic}[1]
        \Require $\mathcal{D}_{\tau}$, $\mathcal{D}_{v}$, $\Im$, $\sigma$, $f$, $b$, $k$
        \State train $f_1(x)$ with random balanced subset\;
        \For{$t$=1 to $k-1$}
          \State $F_{t}(x) = \frac{1}{t}\sum_{i=1}^{t} f_i(x)$\;
          \State compute $\widehat{E}_{\mathcal{D}_{\tau}}$ and $\widehat{E}_{\mathcal{D}_{v}}$ by Eq.~\ref{eq:error-distribution}\;
          \State $s_t = [\widehat{E}_{\mathcal{D}_{\tau}}:\widehat{E}_{\mathcal{D}_{v}}]$\;
          \State $\mu_t \thicksim \Im(\mu_t|s_t)$\;
          \State $\mathcal{D}^{'}_{t\text{+1},\tau} = \text{\tt Sample}(\mathcal{D}_{\tau}; F_{t}, \mu_t, \sigma)$\;
          \State train new classifier $f_{t\text{+1}}(x)$ with $\mathcal{D}^{'}_{t\text{+1},\tau}$\;
        \EndFor
        \State \Return {$F_k(x) = \frac{1}{k}\sum_{i=1}^k f_i(x)$}
      \end{algorithmic}
    \end{algorithm}
  \end{minipage}
\end{table}

{\bf Ensemble Training.}
Given a meta-sampler $\Im: \mathbb{R}^{2b} \to [0,1]$ and the meta-sampling strategy, we can iteratively train new base classifiers using the dataset sampled by the sampler. 
At the $t$-th iteration, having the current ensemble $F_{t}(\cdot)$, we can obtain $\widehat{E}_{\mathcal{D}_{\tau}}$, $\widehat{E}_{\mathcal{D}_{v}}$ and meta-state $s_t$ by applying Eqs. (\ref{eq:error-distribution}) and (\ref{eq:meta-state}). 
Then a new base classifier $f_{t+1}(\cdot)$ is trained with the subset $\mathcal{D}^{'}_{t+1,\tau} = \texttt{Sample}(\mathcal{D}_{\tau}; F_{t}, \mu_t, \sigma)$, where $\mu_t \thicksim \Im(\mu_t|s_t)$ and $\mathcal{D}_{\tau}$ is the original training set. 
Note that $f_1(\cdot)$ was trained on a random balanced subset, as there is no trained classifier in the first iteration. 
See Algorithm~\ref{alg:ensemble-training} for more details. 

\begin{algorithm}[t]
  \caption{Meta-training in {\sc Mesa}}
  \label{alg:meta-training}
  \begin{algorithmic}[1]
  \State Initialization: replay memory $\mathcal{M}$ with capacity $N$, network parameters $\psi, \bar{\psi}, \theta,$ and $\varphi$
  \For {episode = 1 to $M$}
    % \State train base classifier $f_1(x)$ with random balanced subset\;
    \For{each environment step $t$}
      \State observe $s_t$ from $\mathsf{ENV}$ \Comment line3-5 in Alg.~\ref{alg:ensemble-training}\;
      \State take action $\mu_t \thicksim \Im_\varphi(\mu_t|s_t)$ \Comment line6-8 in Alg.~\ref{alg:ensemble-training}\;
      \State observe reward $r_t = P(F_{t\text{+1}}, \mathcal{D}_v) - P(F_{t}, \mathcal{D}_v)$ and $s_{t+1}$
      \State store transition $\mathcal{M} = \mathcal{M} \cup \{(s_t, \mu_t, r_t, s_{t+1})\}$
    \EndFor
    \For{each gradient step}
      \State update $\psi, \bar{\psi}, \theta,$ and $\varphi$ according to~\cite{haarnoja2018soft-actor-critic}
    \EndFor
  \EndFor
  \State \Return {meta-sampler $\Im$ with parameters $\varphi$}
  \end{algorithmic}
\end{algorithm}

{\bf Meta Training.}
As described above, our meta-sampler $\Im$ is trained to optimize the generalized performance of an ensemble classifier by iteratively selecting its training data. 
It takes the current state $s$ of the training system as input, and then outputs the parameter $\mu$ of a Gaussian function to decide each instance's sampling probability. 
The meta-sampler is expected to learn and adapt its strategy from such state($s$)-action($\mu$)-state(new $s$) interactions.
The non-differentiable optimization problem of training $\Im$ can thus be naturally approached via reinforcement learning (RL). 

We consider the ensemble training system as the environment ($\mathsf{ENV}$) in the RL setting. 
The corresponding Markov decision process (MDP) is defined by the tuple ($\mathcal{S}, \mathcal{A}, p, r$), where the state space $\mathcal{S}:\mathbb{R}^{2b}$ and action space $\mathcal{A}:[0,1]$ is continuous, 
and the unknown state transition probability $p: \mathcal{S \times S \times A} \to [0, \infty)$ represents the probability density of the next state $s_{t+1} \in \mathcal{S}$ given the current state $s_t \in \mathcal{S}$ and action $a_t \in \mathcal{A}$. 
More specifically, in each episode, we iteratively train $k$ base classifiers $f(\cdot)$ and form a cascade ensemble classifier $F_k(\cdot)$. 
In each environment step, $\mathsf{ENV}$ provides the meta-state $s_t = [\widehat{E}_{\mathcal{D}_{\tau}}:\widehat{E}_{\mathcal{D}_{v}}]$, and then the action $a_t$ is selected by $a_t \thicksim \Im(\mu_t|s_t)$, i.e., $a_t \Leftrightarrow \mu_t$. 
A new base classifier $f_{t+1}(\cdot)$ is trained using the subset $\mathcal{D}^{'}_{t+1,\tau} = Sample(\mathcal{D}_{\tau}; F_{t}, a_t, \sigma)$. 
After adding $f_{t+1}(\cdot)$ into the ensemble, the new state $s_{t+1}$ was sampled w.r.t. $s_{t+1} \thicksim p(s_{t+1}; s_t, a_t)$. 
Given a performance metric function $P(F, \mathcal{D}) \to \mathbb{R}$, the reward $r$ is set to the generalization performance difference of $F$ before and after an update (using the keep-out validation set for unbiased estimation), i.e., $r_t = P(F_{t+1}, \mathcal{D}_v) - P(F_t, \mathcal{D}_v)$. 
The optimization goal of the meta-sampler (i.e., the cumulative reward) is thus the generalization performance of the ensemble classifier.

We take advantage of Soft Actor-Critic~\cite{haarnoja2018soft-actor-critic} ({\sc Sac}), an off-policy actor-critic deep RL algorithm based on the maximum entropy RL framework, to optimize our meta-sampler $\Im$. 
In our case, we consider a parameterized state value function $V_\psi(s_t)$ and its corresponding target network $V_{\bar{\psi}}(s_t)$, a soft Q-function $Q_\theta(s_t, a_t)$, and a tractable policy (meta-sampler) $\Im_\varphi(a_t|s_t)$. 
The parameters of these networks are $\psi, \bar{\psi}, \theta,$ and $\varphi$. 
The rules for updating these parameters are given in the {\sc Sac} paper~\cite{haarnoja2018soft-actor-critic}. 
We summarize the meta-training process of $\Im_\varphi$ in Algorithm~\ref{alg:meta-training}. 

{\bf Complexity analysis.} 
Please refer to Section~\ref{section:complexity-analysis} for detailed complexity analysis of {\sc Mesa} alongside with related validating experiments in Fig.~\ref{fig:subtask-meta-training-cost}.
% Please refer to the Appendix (provided in supplementary material) for detailed complexity analysis of {\sc Mesa} alongside with related validating experiments.

%% file: sections/experiments.tex
\section{Experiments}
\label{section:experiments}

To thoroughly assess the effectiveness of {\sc Mesa}, two series of experiments are conducted: 
one on controlled synthetic toy datasets for visualization and the other on real-world imbalanced datasets to validate {\sc Mesa}'s performance in practical applications. 
We also carry out extended experiments on real-world datasets to verify the robustness and cross-task transferability of {\sc Mesa}. 

\subsection{Experiment on Synthetic Datasets}

\begin{figure}[t]
  \centering
  \parbox{.5\textwidth}{
  \includegraphics[width=\linewidth]{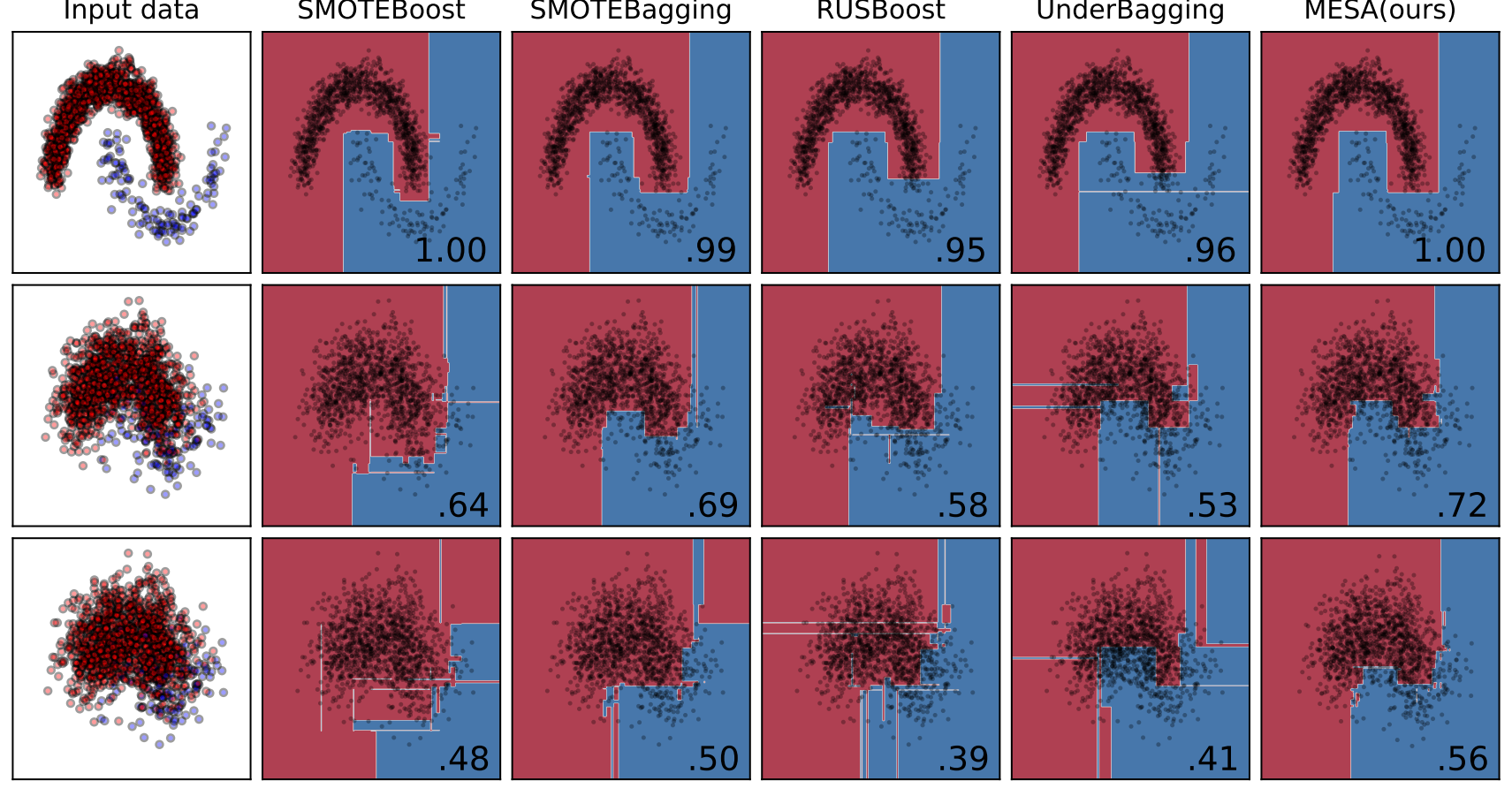}
  }
  \hspace{\fill}
  \parbox{.48\textwidth}{
    \vspace{5pt}
    \caption{
    Comparisons of {\sc Mesa} with 4 representative traditional EIL methods 
    ({\sc SmoteBoost~\cite{chawla2003smoteboost}, SmoteBagging~\cite{wang2009smotebagging}, RusBoost~\cite{seiffert2010rusboost}} and {\sc UnderBagging~\cite{barandela2003underbagging}}) 
    on 3 toy datasets with different levels of underlying class distribution overlapping (less/mid/highly-overlapped in 1st/2nd/3rd row).
    The number in the lower right corner of each subfigure represents the AUCPRC score of the corresponding classifier. 
    Best viewed in color. 
    }
    \label{fig:visualization}
  }
\end{figure}

{\bf Setup Details.}
We build a series of imbalanced toy datasets corresponding to different levels of underlying class distribution overlapping, as shown in Fig.~\ref{fig:visualization}. 
All the datasets have the same imbalance ratio\footnote{Imbalance ratio (IR) is defined as $\mathcal{|N|/|P|}$.} ($|\mathcal{N}|/|\mathcal{P}|=2,000/200=10$). 
In this experiment, {\sc Mesa} is compared with four representative EIL algorithms from 4 major EIL branches (Parallel/Iterative Ensemble + Under/Over-sampling), i.e., {\sc SmoteBoost~\cite{chawla2003smoteboost}, SmoteBagging~\cite{wang2009smotebagging}, RusBoost~\cite{seiffert2010rusboost}}, and {\sc UnderBagging~\cite{barandela2003underbagging}}. 
All EIL methods are deployed with decision trees as base classifiers with ensemble size of $5$.  

{\bf Visualization \& Analysis.}
We plot the input datasets and the decision boundaries learned by different EIL algorithms in Fig.{fig:visualization}, which shows that {\sc Mesa} achieves the best performance under different situations.
We can observe that: all tested methods perform well on the less-overlapped dataset (1st row). 
Note that random under-sampling discards some important majority samples (e.g., data points at the right end of the ``$\cap$''-shaped distribution) and cause information loss. 
This makes the performance of {\sc RusBoost} and {\sc UnderBagging} slightly weaker than their competitors.
As overlapping intensifies (2nd row), an increasing amount of noise gains high sample weights during the training process of boosting-based methods, i.e., {\sc SmoteBoost} and {\sc RusBoost}, thus resulting in poor classification performance. 
Bagging-based methods, i.e., {\sc SmoteBagging} and {\sc UnderBagging}, are less influenced by noise but they still underperform {\sc Mesa}. 
Even on the extremely overlapped dataset (3rd row), {\sc Mesa} still gives a stable and reasonable decision boundary that fits the underlying distribution. 
This is because the meta-sampler can adaptively select informative training subsets towards the good prediction performance while being robust to noise/outliers. 
All the results show the superiority of our {\sc Mesa} to other traditional EIL baselines in handling the distribution overlapping, noises, and poor minority class representation. 

\subsection{Experiment on Real-world Datasets}

{\bf Setup Details.}
In order to verify the effectiveness of {\sc Mesa} in practical applications, we extend the experiments to real-world imbalanced classification tasks from the UCI repository~\cite{Dua2019uci} and KDD CUP 2004. 
To ensure a thorough assessment, these datasets vary widely in their properties, with the imbalance ratio (IR) ranging from 9.1:1 to 111:1, dataset sizes ranging from 531 to 145,751, and number of features ranging from 6 to 617 
(Please see Table~\ref{table:datasets} in Section~\ref{section:implementation-details} for detailed information).
% (Please see Appendix provided in the supplementary material for detailed information).
For each dataset, we keep-out the 20\% validation set and report the result of 4-fold stratified cross-validation (i.e., 60\%/20\%/20\% training/validation/test split).
The performance is evaluated using the area under the precision-recall curve (AUCPRC)\footnote{All results are averaged over 10 independent runs.}, which is an unbiased and more comprehensive metric for class-imbalanced tasks compared to other metrics such as F-score, ROC, and accuracy~\cite{davis2006aucprc}.

\begin{table*}[t]
  \centering
  \tiny
  \caption{
    Comparisons of {\sc Mesa} with other representative resampling methods.
  }
  \label{table:comparison-resampling}
  \begin{tabular}{c|c|ccccc|cc}
  \toprule
  \multirow{2}*{Category} & \multirow{2}*{Method} & \multicolumn{5}{c|}{Protein Homo. (IR=111)} & \#Training & Resampling\\
  \cline{3-7}
  & & KNN & GNB & DT & AdaBoost & GBM & Samples & Time (s)\\
  \hline
  No resampling
  & {\sc - } & 
  0.466 & 0.742 & 0.531 & 0.778 & 0.796 & 87,450 & - \\
  \hline
  \multirow{2}*{Under-sampling} 
  & {\sc RandomUS} & 
  0.146 & 0.738 & 0.071 & 0.698 & 0.756 & 1,554 & 0.068 \\
  & {\sc NearMiss~\cite{mani2003nearmiss}} & 
  0.009 & 0.012 & 0.012 & 0.400 & 0.266 & 1,554 & 3.949 \\
  \hline
  \multirow{5}*{Cleaing-sampling} 
  & {\sc Clean~\cite{laurikkala2001ncr}} & 
  0.469 & 0.744 & 0.488 & 0.781 & 0.811 & 86,196 & 117.739 \\
  & {\sc ENN~\cite{wilson1972enn}} & 
  0.460 & 0.744 & 0.532 & 0.789 & 0.817 & 86,770 & 120.046 \\
  & {\sc TomekLink~\cite{tomek1976tomeklink}} & 
  0.466 & 0.743 & 0.524 & 0.778 & 0.791 & 87,368 & 90.633 \\
  & {\sc AllKNN~\cite{tomek1976allknn}} & 
  0.459 & 0.744 & 0.542 & 0.789 & 0.816 & 86,725 & 327.110 \\
  & {\sc OSS~\cite{kubat1997oss}} & 
  0.466 & 0.743 & 0.536 & 0.778 & 0.789 & 87,146 & 92.234 \\
  \hline
  \multirow{4}*{Over-sampling} 
  & {\sc RandomOS} & 
  0.335 & 0.706 & 0.505 & 0.736 & 0.733 & 173,346 & 0.098 \\
  & {\sc Smote~\cite{chawla2002smote}} & 
  0.189 & 0.753 & 0.304 & 0.700 & 0.719 & 173,346 & 0.576 \\
  & {\sc ADASYN~\cite{he2008adasyn}} & 
  0.171 & 0.679 & 0.315 & 0.717 & 0.693 & 173,366 & 2.855 \\
  & {\sc BorderSmote~\cite{han2005borderline-smote}} & 
  0.327 & 0.743 & 0.448 & 0.795 & 0.711 & 173,346 & 2.751 \\
  \hline
  \multirow{2}*{Over-sampling + Cleaning} 
  & {\sc SmoteENN~\cite{batista2004smoteenn}} & 
  0.156 & 0.750 & 0.308 & 0.711 & 0.750 & 169,797 & 156.641 \\
  & {\sc SmoteTomek~\cite{batista2003smotetomek}} & 
  0.185 & 0.749 & 0.292 & 0.782 & 0.703 & 173,346 & 116.401 \\
  \hline
  \multirow{1}*{Meta-sampler} 
  & {\sc Mesa (OURS, $k$=10)} & 
  {\bf 0.585} & {\bf 0.804} & {\bf 0.832} & {\bf 0.849} & {\bf 0.855} & {\it 1,554$\times$10} & {\it 0.235$\times$10} \\
  \bottomrule
  \end{tabular}
\end{table*}

{\bf Comparison with Resampling Imbalanced Learning (IL) Methods.}
We first compare {\sc Mesa} with resampling techniques, which have been widely used in practice for preprocessing imbalanced data~\cite{haixiang2017learning-from-imb-review}.
We select 12 representative methods from 4 major branches of resampling-based IL, i.e, under/over/cleaning-sampling and over-sampling with cleaning-sampling post-process. 
We test all methods on the challenging highly-imbalanced (IR=111) {\it Protein Homo.} task to check their efficiency and effectiveness.
Five different classifiers, i.e., K-nearest neighbor (KNN), Gaussian Na\"ive Bayes (GNB), decision tree (DT), adaptive boosting (AdaBoost), and gradient boosting machine (GBM), were used to collaborate with different resampling approaches.
We also record the number of samples used for model training and the time used to perform resampling. 

Table~\ref{table:comparison-resampling} details the experiment results.
We show that by learning an adaptive resampling strategy, {\sc Mesa} outperforms other traditional data resampling methods by a large margin while only using a small number of training instances.
In such a highly imbalanced dataset, the minority class is poorly represented and lacks a clear structure. 
Thus over-sampling methods that rely on relations between minority objects (like {\sc Smote}) may deteriorate the classification performance, even though they generate and use a huge number of synthetic samples for training. 
On the other hand, under-sampling methods drop most of the samples according to their rules and results in significant information loss and poor performance.
Cleaning-sampling methods aim to remove noise from the dataset, but the resampling time is considerably high and the improvement is trivial.

\begin{table*}[t]
  \centering
  \tiny
  \caption{
    Comparisons of {\sc Mesa} with other representative under-sampling-based EIL methods.
  }
  \label{table:comparison-under-ensemble}
  \begin{tabular}{c|ccc|ccc|ccc|ccc}
  \toprule
  \multirow{2}*{Method} &
  \multicolumn{3}{c|}{Optical Digits (IR=9.1)} &
  \multicolumn{3}{c|}{Spectrometer (IR=11)} &
  \multicolumn{3}{c|}{ISOLET (IR=12)} &
  \multicolumn{3}{c}{Mammography (IR=42)} \\
  \cline{2-13}
  & $k$=5 & $k$=10 & $k$=20 & $k$=5 & $k$=10 & $k$=20 & $k$=5 & $k$=10 & $k$=20 & $k$=5 & $k$=10 & $k$=20 \\
  \hline
  {\sc RusBoost~\cite{seiffert2010rusboost}} & 
  0.883 & 0.946 & 0.958 & 0.686 & 0.784 & 0.786 & 0.696 & 0.770 & 0.789 & 0.348 & 0.511 & 0.588 \\
  {\sc UnderBagging~\cite{barandela2003underbagging}} & 
  0.876 & 0.927 & 0.954 & 0.610 & 0.689 & 0.743 & 0.688 & 0.768 & 0.812 & 0.307 & 0.401 & 0.483 \\
  {\sc SPE~\cite{liu2019self-paced-ensemble}} & 
  0.906 & 0.959 & 0.969 & 0.688 & 0.777 & 0.803 & 0.755 & 0.841 & 0.895 & 0.413 & 0.559 & 0.664 \\
  {\sc Cascade~\cite{liu2009ee-bc}} & 
  0.862 & 0.932 & 0.958 & 0.599 & 0.754 & 0.789 & 0.684 & 0.819 & 0.891 & 0.404 & 0.575 & 0.670 \\
  \hline
  {\sc Mesa (OURS)} & 
  {\bf 0.929} & {\bf 0.968} & {\bf 0.980} & 
  {\bf 0.723} & {\bf 0.803} & {\bf 0.845} & 
  {\bf 0.787} & {\bf 0.877} & {\bf 0.921} & 
  {\bf 0.515} & {\bf 0.644} & {\bf 0.705} \\
  \bottomrule
  \end{tabular}
\end{table*}

\begin{figure*}[t]
  \centering
  \includegraphics[width=\linewidth]{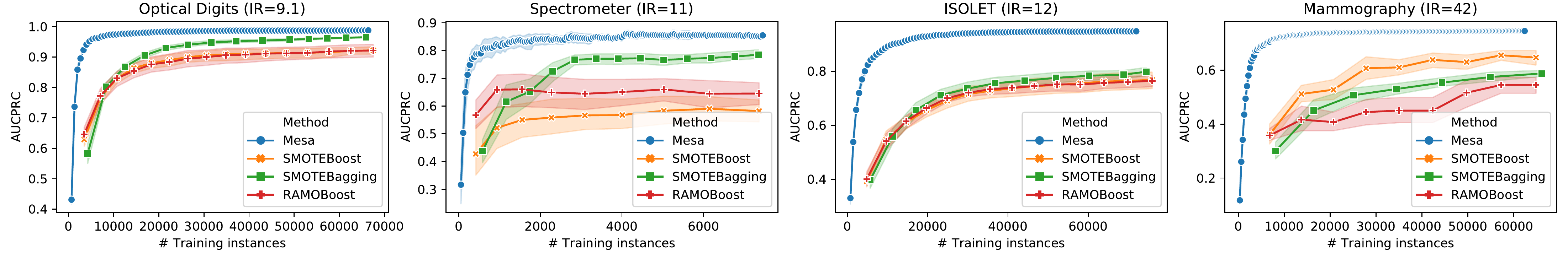}
  \caption{
    Comparisons of {\sc Mesa} with other representative over-sampling-based EIL methods.
  }
  \label{fig:cmp-over-ensemble}
\end{figure*}

{\bf Comparison with Ensemble Imbalanced Learning (EIL) Methods.}
We further compare {\sc Mesa} with 7 representative EIL methods on four real-world imbalanced classification tasks. 
The baselines include 4 under-sampling based EIL methods, i.e., {\sc RusBoost~\cite{seiffert2010rusboost}}, {\sc UnderBagging~\cite{barandela2003underbagging}}, {\sc SPE~\cite{liu2019self-paced-ensemble}}, {\sc Cascade~\cite{liu2009ee-bc}}, and 3 over-sampling-based EIL methods, i.e., {\sc SmoteBoost~\cite{chawla2003smoteboost}, SmoteBagging~\cite{wang2009smotebagging}} and {\sc RamoBoost~\cite{chen2010ramoboost}}.
We use the decision tree as the base learner for all EIL methods following the settings of most of the previous works~\cite{haixiang2017learning-from-imb-review}. 

We report the AUCPRC score of various USB-EIL methods with different ensemble sizes ($k$=5, 10, 20) in Table~\ref{table:comparison-under-ensemble}. 
The results show that {\sc Mesa} achieves competitive performance on various real-world tasks. 
For the baseline methods, we can observe that {\sc RusBoost} and {\sc UnderBagging} suffer from information loss as random under-sampling may discard samples with important information, and such effect is more apparent on highly imbalanced task. 
In comparison, the improved sampling strategies of {\sc SPE} and {\sc Cascade} enable them to achieve relatively better performance but still underperform {\sc Mesa}. 
Moreover, as {\sc Mesa} provides an adaptive resampler that makes the ensemble training converge faster and better, its advantage is particularly evident when using small ensemble in the highly-imbalanced task.
On the {\it Mammography} dataset (IR=42), compared with the second-best score, {\sc Mesa} achieved 24.70\%/12.00\%/5.22\% performance gain when $k$=5/10/20, respectively.

We further compare {\sc Mesa} with 3 OSB-EIL methods.
As summarized in Table~\ref{table:comparison}, OSB-EIL methods typically use much more (1-2$\times$IR times) data to train each base learner than their under-sampling-based competitors, including {\sc Mesa}.
Thus it is unfair to directly compare {\sc Mesa} with over-sampling-based baselines with the same ensemble size. 
Therefore, we plot the performance curve with regard to the number of instances used in ensemble training, as shown in Fig.~\ref{fig:cmp-over-ensemble}.

It can be observed that our method {\sc Mesa} consistently outperforms over-sampling-based methods, especially on highly imbalanced/high-dimensional tasks (e.g., ISOLET with 617 features, Mammo. with IR=42). 
{\sc Mesa} also shows high sample efficiency and faster convergence speed. Compared with the baselines, it only requires a few training instances to converge to a strong ensemble classifier. 
{\sc Mesa} also has a more stable training process. 
The OSB-EIL methods perform resampling by analyzing and reinforcing the structure of minority class data. 
When the dataset is small or highly-imbalanced, the minority class is usually under-represented and lacks a clear structure. 
The performance of these OSB-EIL methods thus becomes unstable under such circumstances.

\begin{table}[t]
  \centering
  \tiny
  \caption{Cross-task transferability of the meta-sampler.}
  \label{table:cross-task-results}
  \begin{tabular}{c|cccc|cccc}
  \toprule
  \multirow{2}*{\diagbox[width=8em]{Meta-train}{Meta-test}} & \multicolumn{4}{c|}{Mammography (IR=42, 11,183 instances)} & \multicolumn{4}{c}{Protein Homo. (IR=111, 145,751 instances)} \\
  \cline{2-9}
  & $k$=10 & $\Delta$ & $k$=20 & $\Delta$ & $k$=10 & $\Delta$ & $k$=20 & $\Delta$ \\
  \hline
  100\%        & 0.644$\pm$0.028 & baseline & 0.705$\pm$0.015 & baseline & 0.840$\pm$0.009 & baseline & 0.874$\pm$0.008 & baseline \\
  50\% subset  & 0.642$\pm$0.032 & -0.30\% & 0.702$\pm$0.017 & -0.43\% & 0.839$\pm$0.009 & -0.12\% & 0.872$\pm$0.009 & -0.23\% \\
  10\% subset  & 0.640$\pm$0.031 & -0.62\% & 0.700$\pm$0.017 & -0.71\% & 0.839$\pm$0.008 & -0.10\% & 0.871$\pm$0.006 & -0.34\% \\
  \hline
  Optical Digits  & 0.637$\pm$0.029 & -1.09\% & 0.701$\pm$0.015 & -0.57\% & 0.839$\pm$0.006 & -0.12\% & 0.870$\pm$0.006 & -0.46\% \\
  Spectrometer    & 0.641$\pm$0.025 & -0.54\% & 0.697$\pm$0.021 & -1.13\% & 0.836$\pm$0.009 & -0.48\% & 0.870$\pm$0.006 & -0.46\% \\
  \bottomrule
  \end{tabular}
\end{table}

{\bf Cross-task Transferability of the Meta-sampler.}
\label{subsubsection:cross-task-transferability}
One important feature of {\sc Mesa} is its cross-task transferability.
As the meta-sampler is trained on task-agnostic meta-data, it is {\it not} task-bounded and is directly applicable to new tasks. 
This provides {\sc Mesa} with better scalability as one can directly use a pre-trained meta-sampler in new tasks thus greatly reduce the meta-training cost.
To validate this, we use {\it Mammography} and {\it Protein Homo.} as two larger and highly-imbalanced meta-test tasks, then consider five meta-training tasks including the original task (baseline), two sub-tasks with {50\%/10\%} of the original training set, and two small tasks {\it Optical Digits} and {\it Spectrometer}. 

Table~\ref{table:cross-task-results} reports the detailed results.
We can observe that the transferred meta-samplers generalize well on meta-test tasks. 
Scaling down the number of meta-training instances has a minor effect on the obtained meta-sampler, especially when the original task has a sufficient number of training samples (e.g., for {\it Protein Homo.}, reducing the meta-training set to 10\% subset only results in -0.10\%/-0.34\% $\Delta$ when $k$=10/20).
Moreover, the meta-sampler that trained on a small task also demonstrates noticeably satisfactory performance (superior to other baselines) on new, larger, and even heterogeneous tasks, which validates the generality of the proposed {\sc Mesa} framework.
Please refer to Section~\ref{section:additional-results} for a comprehensive cross/sub-task transferability test and other additional experimental results.
% Please refer to the Appendix for a comprehensive cross/sub-task transferability test and other additional experimental results.

%% file: sections/conclusion.tex
\section{Conclusion}
\label{section:conclusion}

We propose a novel imbalanced learning framework {\sc Mesa}. 
It contains a meta-sampler that adaptively selects training data to learn effective cascade ensemble classifiers from imbalanced data. 
Rather than following random heuristics, {\sc Mesa} directly optimizes its sampling strategy for better generalization performance. 
Compared with prevailing meta-learning IL solutions that are limited to be co-optimized with DNNs, {\sc Mesa} is a generic framework capable of working with various learning models. 
Our meta-sampler is trained over task-agnostic meta-data and thus can be transferred to new tasks, which greatly reduces the meta-training cost. 
Empirical results show that {\sc Mesa} achieves superior performance on various tasks with high sample efficiency.
In future work, we plan to explore the potential of meta-knowledge-driven ensemble learning in the long-tail multi-classification problem. 

%% file: sections/broader-impact.tex
\section{Statement of the Potential Broader Impact}

In this work, we study the problem of {\it imbalanced learning} (IL), which is a common problem related to machine learning and data mining. 
Such a problem widely exists in many real-world application domains such as finance, security, biomedical engineering, industrial manufacturing, and information technology~\cite{haixiang2017learning-from-imb-review}.
IL methods, including the proposed {\sc Mesa} framework in this paper, aim to fix the bias of learning models introduced by skewed training class distribution. 
We believe that proper usage of these techniques will lead us to a better society. 
For example, better IL techniques can detect phishing websites/fraud transactions to protect people's property, and help doctors diagnose rare diseases/develop new medicines to save people's lives.
With that being said, we are also aware that using these techniques improperly can cause negative impacts, as misclassification is inevitable in most of the learning systems.
In particular, we note that when deploying IL systems in medical-related domains, misclassification (e.g., failure to identify a patient) could lead to medical malpractice.
In such domains, these techniques should be used as auxiliary systems, e.g., when performing diagnosis, we can adjust the classification threshold to achieve higher recall and use the predicted probability as a reference for the doctor's diagnosis.
While there are some risks with IL research, as we mentioned above, we believe that with proper usage and monitoring, the negative impact of misclassification could be minimized and IL techniques can help people live a better life.

%% file: sections/reproducibility.tex
\newpage

\section{Additional Results}
\label{section:additional-results}

\subsection{Cross-task and sub-task transferability of the meta-sampler}
\label{section:additional-results-transfer}

\begin{figure}[h]
  \centering
  \subfigure[Cross-task transfer performance.]{
		\centering
		\includegraphics[width=0.48\linewidth]{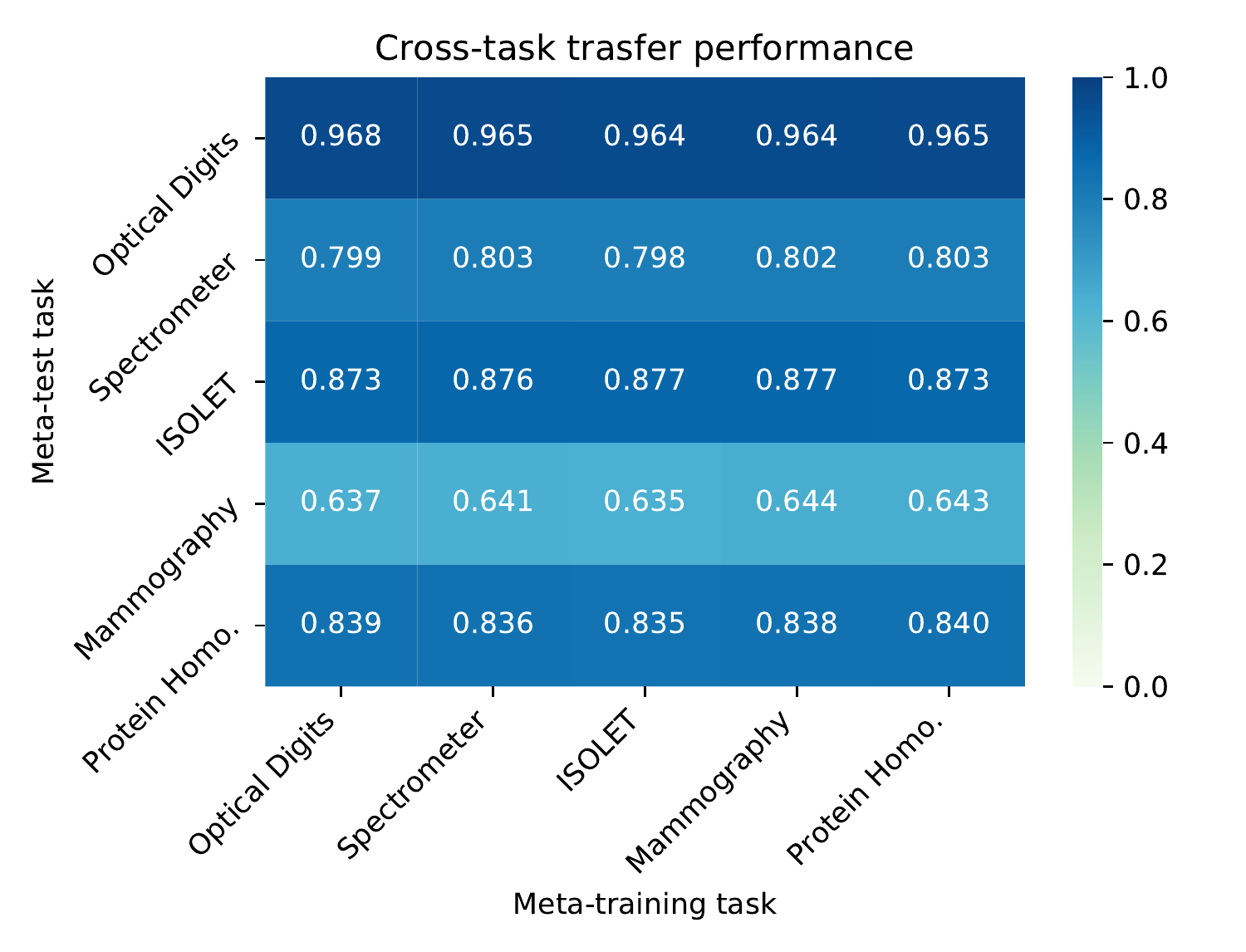}
		\label{fig:transfer-cross-task-performance}
  }
  \subfigure[Cross-task transfer performance loss.]{
		\centering
		\includegraphics[width=0.48\linewidth]{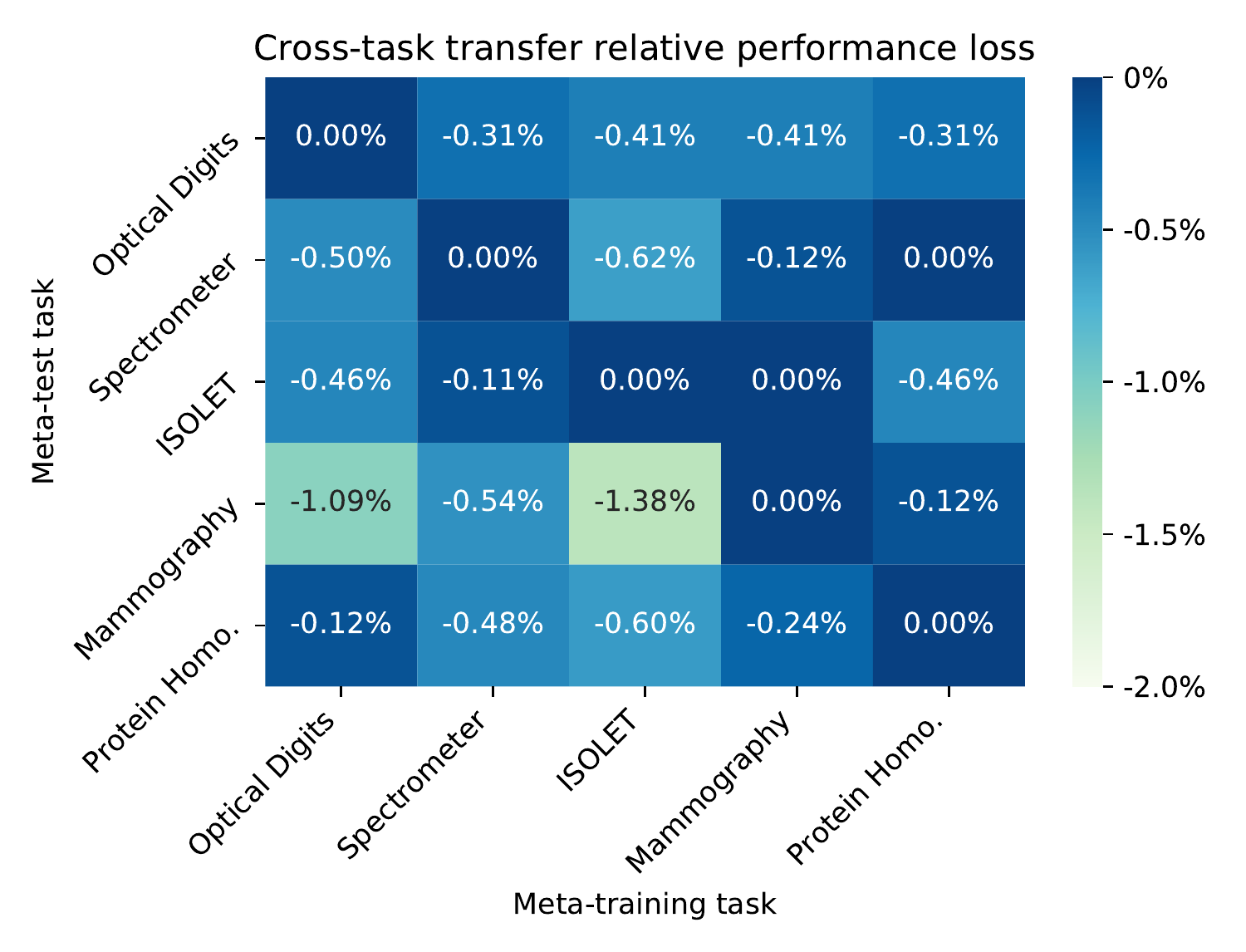}
		\label{fig:transfer-cross-task-performance-loss}
  }
  \subfigure[Sub-task transfer performance.]{
		\centering
		\includegraphics[width=0.48\linewidth]{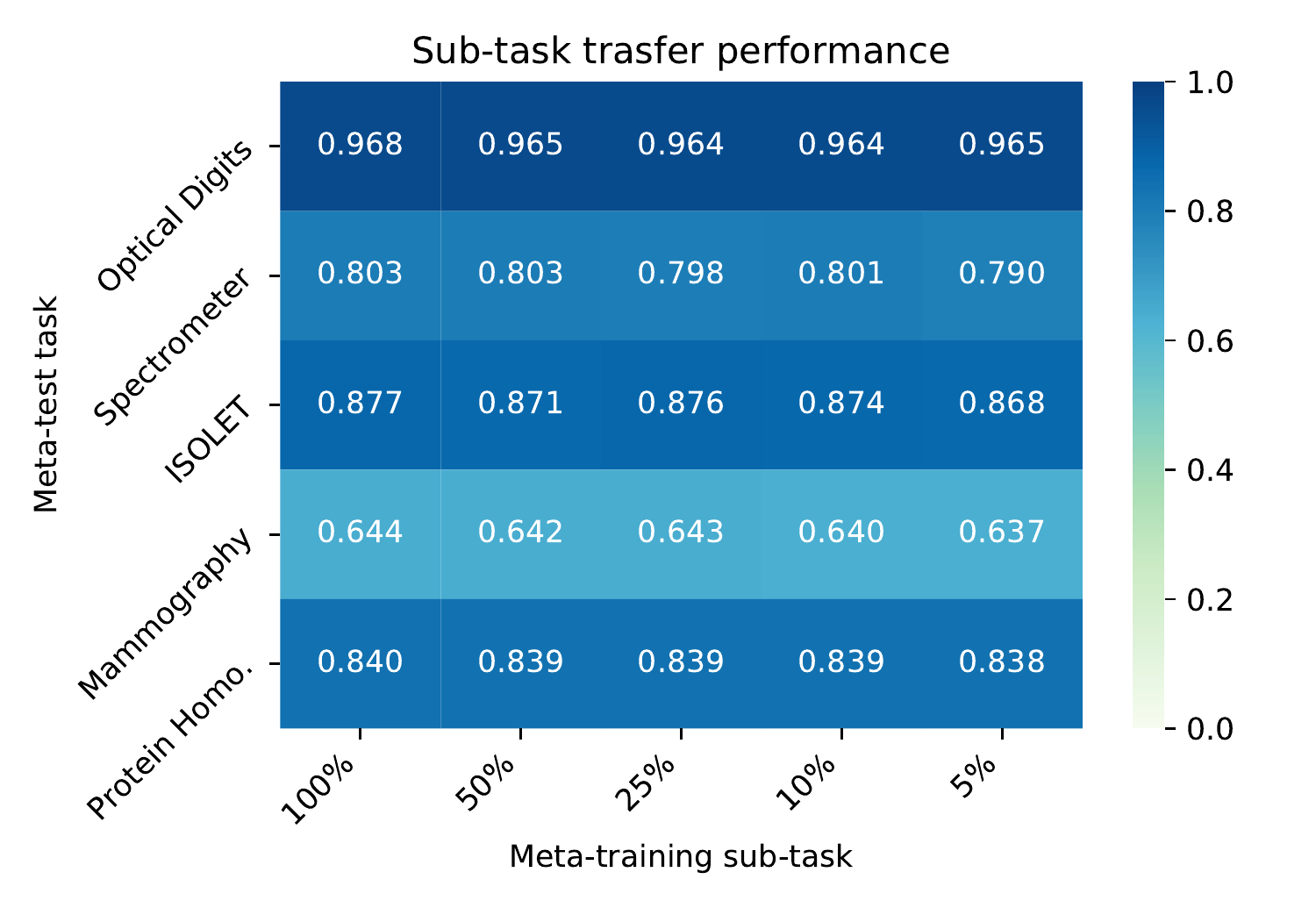}
		\label{fig:transfer-sub-task-performance}
  }
  \subfigure[Sub-task transfer performance loss.]{
		\centering
		\includegraphics[width=0.48\linewidth]{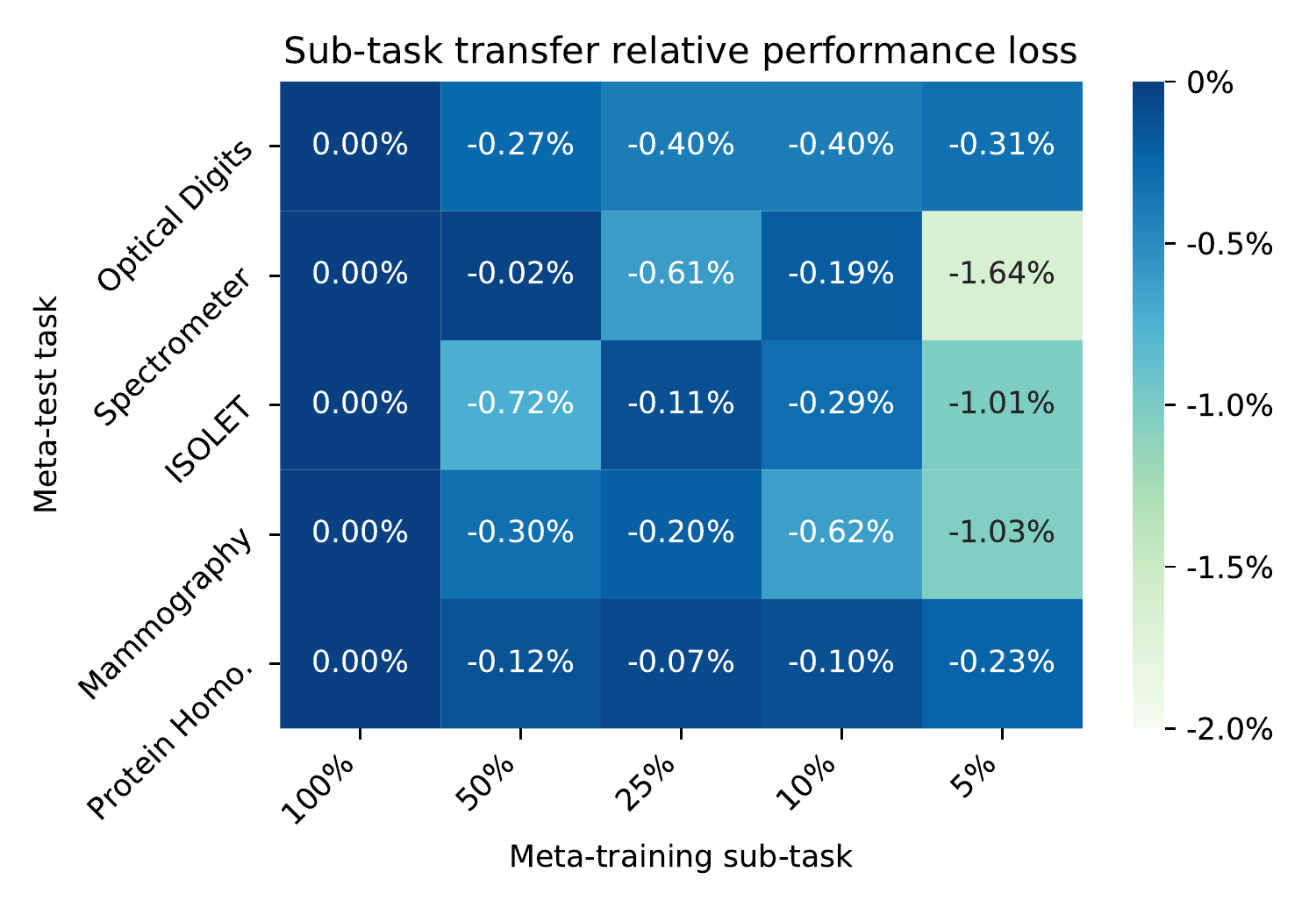}
		\label{fig:transfer-sub-task-performance-loss}
  }
  \caption{Cross/Sub-task transfer performance loss of {\sc Mesa}.}
  \label{fig:transfer-heatmap}
\end{figure}

In addition to results reported in Table~\ref{table:cross-task-results}, we conduct further experiments on all five tasks to test the cross-task transferability of the meta-sampler.
The results are presented in Fig.~\ref{fig:transfer-heatmap} (with $k$=10).
For the cross-task transfer experiment, we meta-train the meta-sampler on each task separately, then apply it on other unseen meta-test tasks. 
As shown in Fig.~\ref{fig:transfer-cross-task-performance-loss}, in all 20 heterogenous training-test task pairs, 18/20 of them manage to have less than 1\% performance loss.
On the other hand, in the sub-task transfer experiment, for each task, we meta-train the meta-sampler on 100\%/50\%/25\%/10\%/5\% subset, then apply it back to the original full dataset. 
Again {\sc Mesa} shows robust performance, in all 20 subset transfer experiments, 17/20 of them manage to have less than 1\% performance loss. 
The effect of reducing the meta-training set scale is more significant in small datasets. The largest performance loss (-1.64\%) is reported in \{5\%, {\it Spectrometer}\} setting, which is the smallest dataset with only 531 instances.
For large datasets, scaling down the meta-training set greatly reduces the number of instances as well as meta-training costs, while only brought about minor performance loss, e.g., -0.23\% loss in \{5\%, {\it Protein Homo.}\}.

\subsection{Robustness to corrupted labels.}
In practice, the collected training dataset may contain corrupted labels. 
Typical examples include data labeled by crowdsourcing systems or search engines~\cite{hendrycks2018using-trusted-data-noisy-labels,li2017learning-from-noisy-labels}. 
The negative impact brought by noise is particularly prominent on skewed datasets that inherently have an unclear minority data structure. 
In this experiment, {\it Mammography} and {\it Protein Homo.} tasks are used to test the robustness of different EIL methods on highly-imbalanced datasets.
We simulate real-world corrupted labels by introducing flip noise. 
Specifically, flip the labels of $r_{\text{noise}}$\% (i.e., $|\mathcal{P}| \cdot r_{\text{noise}}$) minority samples in the training set from 1 to 0.
Accordingly, an equal number of majority samples are flipped from 0 to 1. 
We thereby get a noisy dataset with the same IR. 
For each dataset, we test the USB-EIL methods with $k=10$ trained on the 0\%/10\%/25\%/40\%\ noisy training sets. 

The results are summarized in Table~\ref{table:label-noise}, which shows that {\sc Mesa} consistently outperforms other baselines under different levels of label noise. 
The meta-sampler $\Im$ in {\sc Mesa} can efficiently prevent the ensemble classifier from overfitting noise as it is optimized for generalized performance, while the performance of other methods decrease rapidly as the noise level increases. 
Compared with the second-best baselines, {\sc Mesa} achieves 12.00\%/14.44\%/10.29\%/7.22\% ({\it Mammography}) and 2.56\%/3.20\%/6.92\%/9.72\% ({\it Protein Homo.}) performance gain when $r_{\text{noise}}$=0\%/10\%/25\%/40\%.

\begin{table*}[t]
  \centering
  \tiny
  \caption{Generalized performances on real-world imbalanced datasets with varying label noise ratios.}
  \label{table:label-noise}
  \begin{tabular}{c|cccc|cccc}
  \toprule
  \multirow{2}*{\diagbox[width=8em]{Method}{Dataset}} & 
  \multicolumn{4}{c|}{Mammography (IR=42, 11,183 instances)} &
  \multicolumn{4}{c}{Protein Homo. (IR=111, 145,751 instances)} \\
  \cline{2-9}
  & $r_{\text{noise}}$=0\% & $r_{\text{noise}}$=10\% & $r_{\text{noise}}$=25\% & $r_{\text{noise}}$=40\% & $r_{\text{noise}}$=0\% & $r_{\text{noise}}$=10\% & $r_{\text{noise}}$=25\% & $r_{\text{noise}}$=40\% \\
  \hline
  {\sc RusBoost~\cite{seiffert2010rusboost}} & 
  0.511 & 0.448 & 0.435 & 0.374 & 0.738 & 0.691 & 0.628 & 0.502 \\
  {\sc UnderBagging~\cite{barandela2003underbagging}} & 
  0.401 & 0.401 & 0.375 & 0.324 & 0.632 & 0.629 & 0.629 & 0.617 \\
  {\sc SPE~\cite{liu2019self-paced-ensemble}} & 
  0.559 & 0.476 & 0.405 & 0.345 & 0.819 & 0.775 & 0.688 & 0.580 \\
  {\sc Cascade~\cite{liu2009ee-bc}} & 
  0.575 & 0.540 & 0.447 & 0.357 & 0.805 & 0.781 & 0.708 & 0.594 \\
  \hline
  {\sc Mesa (OURS)} & 
  {\bf 0.644} & {\bf 0.618} & {\bf 0.493} & {\bf 0.401} & {\bf 0.840} & {\bf 0.806} & {\bf 0.757} & {\bf 0.677} \\
  \bottomrule
  \end{tabular}
\end{table*}

\subsection{Cross-task meta-training}
\begin{figure*}[t]
  \centering
  \setlength{\fboxrule}{0.5pt}
  \setlength{\fboxsep}{0pt}
  \shadowsize = 2pt
  \subfigure[Performance in meta-training tasks]{
    \doublebox{\includegraphics[width=0.471\linewidth]{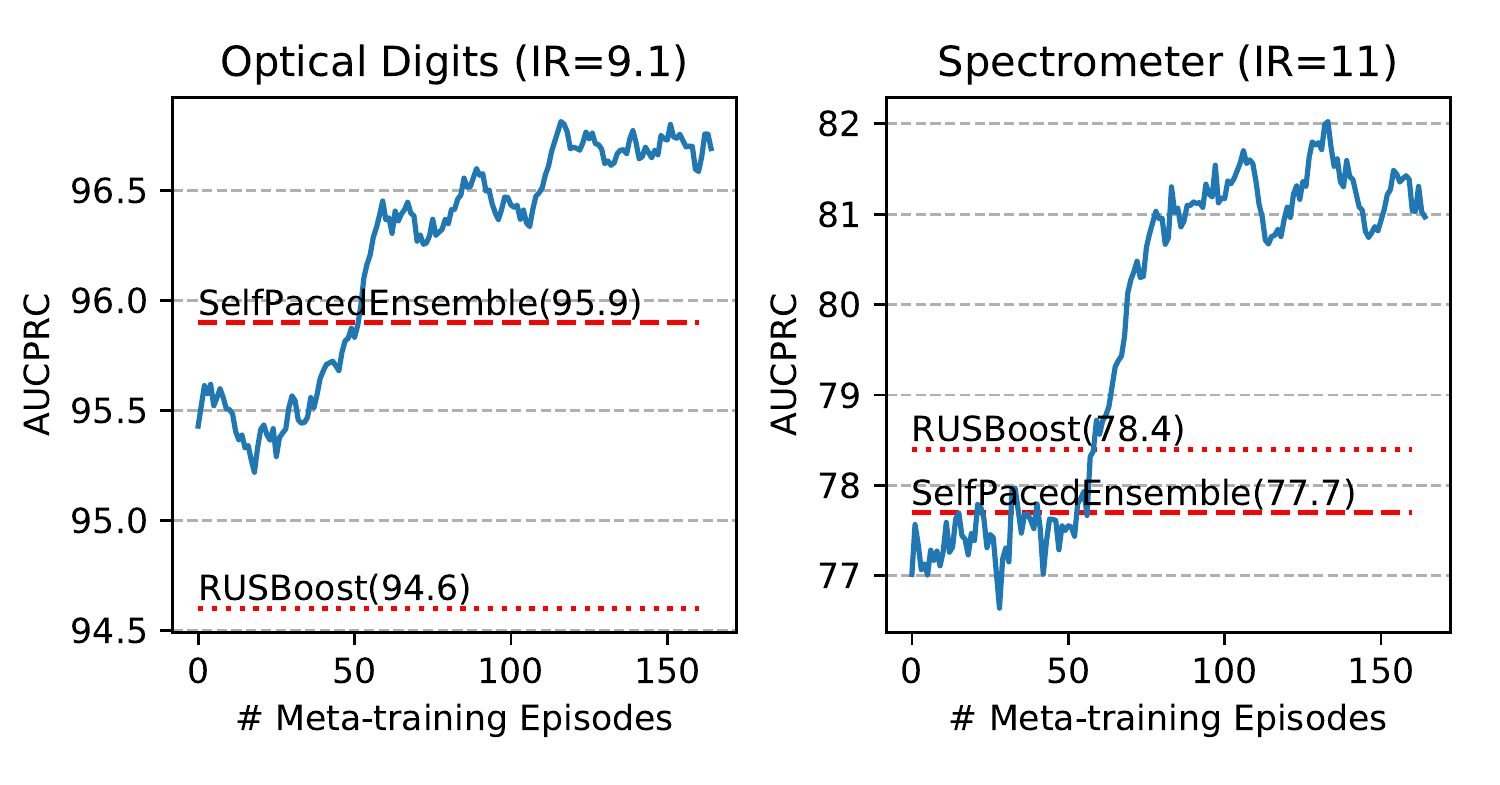}}
    \label{fig:meta-training-tasks}
  }
  \subfigure[Performance in meta-test tasks]{
    \doublebox{\includegraphics[width=0.471\linewidth]{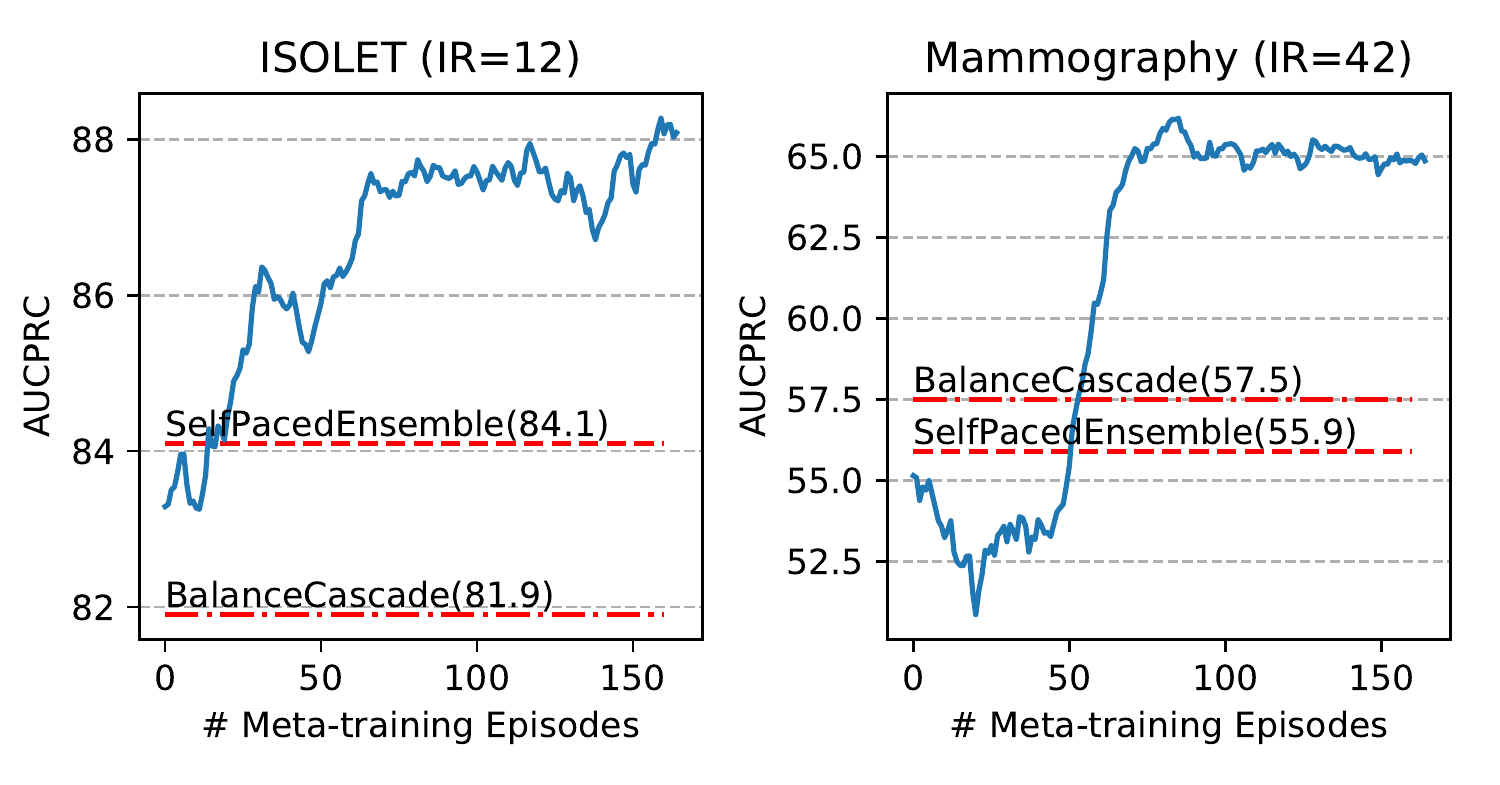}}
    \label{fig:meta-test-tasks}
  }
  \caption{
    Visualization of {\sc Mesa}'s cross-task meta-training process (slide mean window = 50).
  }
  \label{fig:meta-training-process}
\end{figure*}

In the meta-training process of {\sc Mesa}, collecting transitions is independent of the updates of the meta-sampler. 
This enables us to simultaneously collect meta-data from multiple datasets and thus to co-optimize the meta-sampler over these tasks. 
There may be some states that can rarely be observed in a specific dataset, in such case, parallelly collecting transitions from multiple datasets also helps our meta-sampler exploring the state space and learns a better policy. 
Moreover, as previously discussed, a converged meta-sampler can be directly applied to new and even heterogeneous tasks. 
Hence by cross-task meta-training, we can obtain a meta-sampler that not only works well on training tasks but is also able to boost {\sc Mesa}'s performance on unseen (meta-test) tasks. 
To verify this, we follow the setup in section~\ref{subsubsection:cross-task-transferability} using two small tasks for cross-task meta-training and two large tasks for the meta-test. 
We plot the generalized performance on all the four tasks during the cross-task meta-training process, as shown in Fig~\ref{fig:meta-training-process}. 
The performance scores of other representative EIL methods from Table~\ref{table:comparison-under-ensemble} are also included. 
Note that we only plot the two best performing baselines in each subfigure for better visualization. 

At the very start of meta-training, the meta-sampler $\Im$ is initialized with random weights. 
Its performance is relatively poor at this point. 
But as meta-training progresses, $\Im$ adjusts its sampling strategy to maximize the expected generalized performance. 
After 50-60 training episodes, {\sc Mesa} surpasses the best performing baseline method and continued to improve. 
Finally, we get a meta-sampler that is able to undertake adaptive under-sampling and thereby outperform other EIL methods on all meta-training and meta-test tasks. 

\subsection{Ablation study}

To assess the importance of Gaussian function weighted meta-sampling and meta-sampler respectively, we carry out ablation experiments on 4 real-world datasets. 
They are Optical Digits, Spectrometer, ISOLET, and Mammography with increasing IR (9.1/11/12/42).
Our experiments shown in Table~\ref{table:ablation-test-mesa} indicate that {\sc Mesa} significantly improves performance, especially when using small ensembles on highly imbalanced datasets. 

\begin{table}[t]
  % \scriptsize
  \small
  \centering
  \caption{
    Ablation study of {\sc Mesa} on 4 real-world datasets. 
    Random policy refers to using randomly initialized meta-sampler to perform meta-sampling.
    $k$ represents the ensemble size. 
    $\Delta$ is the relative performance loss (\%) compared to {\sc Mesa} policy. 
    }
  \label{table:ablation-test-mesa}
  \begin{tabular}{c|c|cccccc}
  \toprule
  Dataset & Method & $k=5$ & $\Delta$ & $k=10$ & $\Delta$ & $k=20$ & $\Delta$ \\
  \midrule
  \multirow{3}*{Optical Digits} 
  & {\sc Mesa} policy & 0.929 & baseline & 0.968 & baseline & 0.980 & baseline \\
  & Random policy     & 0.904 & -1.61\%  & 0.959 & -0.93\%  & 0.975 & -0.51\% \\
  & Random sampling   & 0.876 & -5.71\%  & 0.927 & -4.24\%  & 0.954 & -2.65\% \\
  \midrule
  \multirow{3}*{Spectrometer} 
  & {\sc Mesa} policy & 0.723 & baseline & 0.803 & baseline & 0.845 & baseline \\
  & Random policy     & 0.685 & -5.26\%  & 0.774 & -3.61\%  & 0.800 & -5.33\% \\
  & Random sampling   & 0.610 & -15.63\% & 0.692 & -13.82\% & 0.755 & -10.65\%\\
  \midrule
  \multirow{3}*{ISOLET} 
  & {\sc Mesa} policy & 0.787 & baseline & 0.877 & baseline & 0.921 & baseline \\
  & Random policy     & 0.748 & -4.96\%  & 0.849 & -3.19\%  & 0.891 & -3.26\%\\
  & Random sampling   & 0.688 & -12.58\% & 0.768 & -12.43\% & 0.812 & -11.83\%\\
  \midrule
  \multirow{3}*{Mammography} 
  & {\sc Mesa} policy & 0.515 & baseline & 0.644 & baseline & 0.705 & baseline \\
  & Random policy     & 0.405 & -21.36\% & 0.568 & -11.80\% & 0.662 & -6.10\%\\
  & Random sampling   & 0.307 & -40.39\% & 0.401 & -37.73\% & 0.483 & -31.49\%\\
  \bottomrule
  \end{tabular}
\end{table}

\section{Implementation Details}
\label{section:implementation-details}

\begin{table*}[t]
  \centering
  % \scriptsize
  \small
  \caption{Description of the real-world imbalanced datasets.}
  \label{table:datasets}
  \begin{tabular}{c|cc|ccc}
  \toprule
  Dataset & Repository & Target & Imbalance Ratio & \#Samples & \#Features\\
  \midrule
  Optical Digits & UCI & target: 8 & 9.1:1 & 5,620 & 64 \\
  Spectrometer   & UCI & target: $\ge 44$ & 11:1 & 531 & 93 \\
  ISOLET         & UCI & target: A, B & 12:1 & 7,797 & 617 \\
  Mammography    & UCI & target: minority & 42:1 & 11,183 & 6 \\
  Protein Homo.  & KDDCUP 2004 & target: minority & 111:1 & 145,751 & 74 \\
  \bottomrule
  \end{tabular}
\end{table*}

{\bf Datasets.}
All datasets used in this paper are publicly available, and are summarized in Table~\ref{table:datasets}.
One can fetch these datasets using the {\tt imblearn.dataset} API\footnote{\tt https://imbalanced-learn.readthedocs.io/en/stable/api.html} of the imbalanced-learn~\cite{guillaume2017imblearn} Python package. 
For each dataset, we keep-out the 20\% validation set and report the result of 4-fold stratified cross-validation (i.e., 60\%/20\%/20\% train/valid/test split).
We also perform class-wise split to ensure that the imbalanced ratio of the training, validation, and test sets after splitting is the same.

{\bf Base classifiers.}
All used base classifiers (i.e., K-nearest neighbor classifier, Gaussian native bayes, decision tree, adaptive boosting, gradient boosting machine) are implemented using {\tt scikit-learn}~\cite{pedregosa2011sklearn} Python package.
For the ensemble models (i.e., adaptive boosting and gradient boosting), we set the {\tt n\_estimators} = 10.
All other parameters use the default setting specified by the {\tt scikit-learn} package.

{\bf Implementation of baseline methods.}
All baseline resampling IL methods 
({\sc RandomUS, NearMiss~\cite{mani2003nearmiss}, Clean~\cite{laurikkala2001ncr}, ENN~\cite{wilson1972enn}, TomekLink~\cite{tomek1976tomeklink}, AllKNN~\cite{tomek1976allknn}, OSS~\cite{kubat1997oss}, Smote~\cite{chawla2002smote}, ADASYN~\cite{he2008adasyn}, BorderSmote~\cite{han2005borderline-smote}, SmoteENN~\cite{batista2004smoteenn},} and {\sc SmoteTomek~\cite{batista2003smotetomek}}) 
are implemented in {\tt imbalanced-learn} Python package~\cite{guillaume2017imblearn}. 
We directly use their implementation and default hyper-parameters in our experiments.
We use open-source code\footnote{\tt https://github.com/dialnd/imbalanced-algorithms}\footnote{\tt https://github.com/ZhiningLiu1998/self-paced-ensemble} for implementation of baseline ensemble imbalanced learning (EIL) methods ({\sc RusBoost~\cite{seiffert2010rusboost}, UnderBagging~\cite{barandela2003underbagging}, Cascade~\cite{liu2009ee-bc}, SPE~\cite{liu2019self-paced-ensemble}, SmoteBoost~\cite{chawla2003smoteboost}, SmoteBagging~\cite{wang2009smotebagging},} and {\sc RamoBoost~\cite{chen2010ramoboost}}).
The hyper-parameters of these baseline EIL methods are reported in Table~\ref{table:hyper-parameters-baseline}.

{\bf Implementation of {\sc Mesa}.}
{\sc Mesa} is implemented with {\tt PyTorch}. 
The empirical results reported in the paper use hyper-parameters in Tables~\ref{table:hyper-parameters-sac} and~\ref{table:hyper-parameters-mesa} for the meta-training of {\sc Mesa}. 
We open-sourced our {\sc Mesa} implementation at Github\footnote{\codeurl} with a {\it jupyter notebook} file that allows you to quickly (I) conduct a comparative experiment, (II) visualize the meta-training process of {\sc Mesa}, and (III) visualize the experimental results.
Please check the repository for more information.

\begin{table}[t]
  \scriptsize
  \begin{minipage}{0.48\linewidth}
    \centering
    \caption{Hyper-parameters of EIL baselines.}
    \label{table:hyper-parameters-baseline}
    \begin{tabular}{c|c|c}
    \toprule
    Method & Hyper-parameter & Value \\
    \midrule
    \multirow{5}*{\sc RusBoost~\cite{seiffert2010rusboost}}
    & n\_samples & 100 \\
    & min\_ratio & 1.0 \\
    & with\_replacement & True \\
    & learning\_rate & 1.0 \\
    & algorithm & SAMME.R \\
    \midrule
    \multirow{4}*{\sc SmoteBoost~\cite{chawla2003smoteboost}}
    & n\_samples & 100 \\
    & k\_neighbors & 5 \\
    & learning\_rate & 1.0 \\
    & algorithm & SAMME.R \\
    \midrule
    \multirow{6}*{\sc RamoBoost~\cite{chen2010ramoboost}}
    & n\_samples & 100 \\
    & k\_neighbors\_1 & 5 \\
    & k\_neighbors\_2 & 5 \\
    & alpha & 0.3 \\
    & learning\_rate & 1.0 \\
    & algorithm & SAMME.R \\
    \midrule
    {\sc UnderBagging~\cite{barandela2003underbagging}} & - - - & - - - \\
    \midrule
    {\sc SmoteBagging~\cite{wang2009smotebagging}} & k\_neighbors & 5 \\
    \midrule
    {\sc BalanceCascade~\cite{liu2009ee-bc}} & - - - & - - - \\
    \midrule
    \multirow{2}*{\sc SelfPacedEnsemble~\cite{liu2019self-paced-ensemble}}
    & hardness\_func & cross entropy \\
    & k\_bins & 10 \\
    \bottomrule
    \end{tabular}
  \end{minipage}
  \hspace{5pt}
  \begin{minipage}{0.5\linewidth}
    \centering
    \small
    \caption{Hyper-parameters of {\sc SAC~\cite{haarnoja2018soft-actor-critic}}.}
    \label{table:hyper-parameters-sac}
    \begin{tabular}{c|c}
    \toprule
    Hyper-parameter & Value \\
    \midrule
    Policy type & Gaussian \\
    Reward discount factor ($\gamma$) & 0.99 \\
    Smoothing coefficient ($\tau$) & 0.01 \\
    Temperature parameter ($\alpha$) & 0.1 \\
    Learning rate & 1e-3 \\
    Learning rate decay steps & 10 \\
    Learning rate decay ratio & 0.99 \\
    Mini-batch size & 64 \\
    Replay memory size & 1e3 \\
    Steps of gradient updates & 1e3 \\
    Steps of random actions & 5e2 \\
    % Value target update per no. of updates per step & 1 \\
    \bottomrule
    \end{tabular}

    \vspace{18pt}

    \centering
    \caption{Hyper-parameters of {\sc Mesa}.}
    \label{table:hyper-parameters-mesa}
    \begin{tabular}{c|c}
    \toprule
    Hyper-parameter & Value \\
    \midrule
    Meta-state size & 10 \\
    % Ensemble size $k$ in meta-training & 10 \\
    Gaussian function parameter $\sigma$ & 0.2 \\
    \bottomrule
    \end{tabular}
  \end{minipage}
\end{table}

The actor policy network of meta-sampler is a multi-layer perceptron with one hidden layer containing 50 nodes. 
Its architecture is thus \{{\tt state\_size}, 50, 1\}. 
The corresponding (target) critic Q-network is also an MLP but with two hidden layers. 
As it takes both state and action as input, its architecture is thus \{{\tt state\_size}+1, 50, 50, 1\}. 
Each hidden node is with ReLU activation function, and the output of the policy network is with the tanh activation function, to guarantee the output located in the interval of $[0,1]$. 
As a general network training trick, we employ the Adam optimizer to optimize the policy and critic networks. 

\begin{table}[t]
  % \small
  % \scriptsize
  \footnotesize
  \centering
  \caption{Performance of different policy network architectures.}
  \label{table:policy-network-structures-mesa}
  \begin{tabular}{c|ccc}
  \toprule
  \multirow{2}*{Network Architecture} & \multicolumn{3}{c}{Optical Digits Task} \\
  \cline{2-4}
  & $k$=5 & $k$=10 & $k$=20 \\
  \midrule
  \{10, 50, 1\}     & 0.929$\pm$0.015 & 0.968$\pm$0.007 & 0.980$\pm$0.003 \\
  \{10, 100, 1\}    & 0.930$\pm$0.014 & 0.966$\pm$0.007 & 0.979$\pm$0.004 \\
  \{10, 200, 1\}    & 0.922$\pm$0.018 & 0.964$\pm$0.008 & 0.978$\pm$0.005 \\
  \{10, 25, 25, 1\} & 0.928$\pm$0.014 & 0.966$\pm$0.007 & 0.980$\pm$0.004 \\
  \{10, 50, 50, 1\} & 0.929$\pm$0.017 & 0.967$\pm$0.008 & 0.978$\pm$0.004 \\
  \{10, 100, 100, 1\}   & 0.926$\pm$0.015 & 0.966$\pm$0.010 & 0.979$\pm$0.006 \\
  \{10, 10, 10, 10, 1\} & 0.924$\pm$0.013 & 0.964$\pm$0.007 & 0.977$\pm$0.004 \\
  \{10, 25, 25, 25, 1\} & 0.924$\pm$0.016 & 0.966$\pm$0.006 & 0.978$\pm$0.002 \\
  \{10, 50, 50, 50, 1\} & 0.926$\pm$0.006 & 0.965$\pm$0.006 & 0.979$\pm$0.005 \\
  \bottomrule
  \end{tabular}
\end{table}

We test different network architecture settings in experiments. 
Table~\ref{table:policy-network-structures-mesa} depicts some representative results under 9 different policy network structures, with different depths and widths. 
It can be observed that varying MLP settings have no substantial effects on the final result. 
We hence prefer to use the simple and shallow one. 

\section{Discussion}

\subsection{Complexity analysis of the proposed framework}
\label{section:complexity-analysis}

Our {\sc Mesa} framework can be roughly regarded as an under-sampling-based ensemble imbalanced learning (EIL) framework (Algorithm~\ref{alg:ensemble-training}) with an additional sampler meta-training process (Algorithm~\ref{alg:meta-training}).

{\bf Ensemble training.}
Given an imbalanced dataset $\mathcal{D}$ with majority set $\mathcal{N}$ and minority set $\mathcal{P}$, where $|\mathcal{N}| \gg |\mathcal{P}|$.
Suppose that the cost of training a base classifier $f(\cdot)$ with $N$ training instances is $C_{f\text{train}}(N)$.
As {\sc Mesa} performs strictly balanced under-sampling to train each classifier, 
% the cost of training a $k$-classifier ensemble is $k \cdot C_{f\text{train}}(2|\mathcal{P}|)$.
we have 
\[
  \text{Cost of $k$-classifier ensemble training}: k \cdot C_{f\text{train}}(2|\mathcal{P}|)
\]
In comparison, the cost is $k \cdot C_{f\text{train}}(|\mathcal{N}|+|\mathcal{P}|)$ for reweighting-based EIL methods (e.g., {\sc AdaBoost}) and around $k \cdot C_{f\text{train}}(2|\mathcal{N}|)$ for over-sampling-based EIL methods (e.g., {\sc SmoteBagging}).

{\bf Meta-training.}
Let's denote the cost of performing a single gradient update step of the meta-sampler $\Im$ as $C_{\Im \text{update}}$, this cost mainly depends on the choice of the policy/critic network architecture. 
It is barely influenced by other factors such as the dataset size in ensemble training. 
In our {\sc Mesa} implementation, we do $n_\text{random}$ steps for collecting transitions with random actions before start updating $\Im$, and $n_\text{update}$ steps for collecting online transitions and perform gradient updates to $\Im$.
% Then the cost of meta-training can be written as $(n_\text{random}+n_\text{update}) \cdot C_{f\text{train}}(2|\mathcal{P}|) + n_\text{update} \cdot C_{\Im \text{update}}$
Then we have 
\[
  \text{Cost of meta-training}: (n_\text{random}+n_\text{update}) \cdot C_{f\text{train}}(2|\mathcal{P}|) + n_\text{update} \cdot C_{\Im \text{update}}
\]
As mentioned before, the meta-training cost can be effectively reduced by scaling down the meta-training dataset (i.e., reducing $|\mathcal{P}|$).
This can be achieved by using a subset of the original data in meta-training. 
One can also directly use a meta-sampler pre-trained on other (smaller) dataset to avoid the meta-training phase when applying {\sc Mesa} to new tasks. 
Both ways should only bring minor performance loss, as reported in Fig.~\ref{fig:transfer-heatmap}.

\begin{figure}[t]
  \centering
  \subfigure[Sub-task transfer performance.]{
		\centering
		\includegraphics[width=0.48\linewidth]{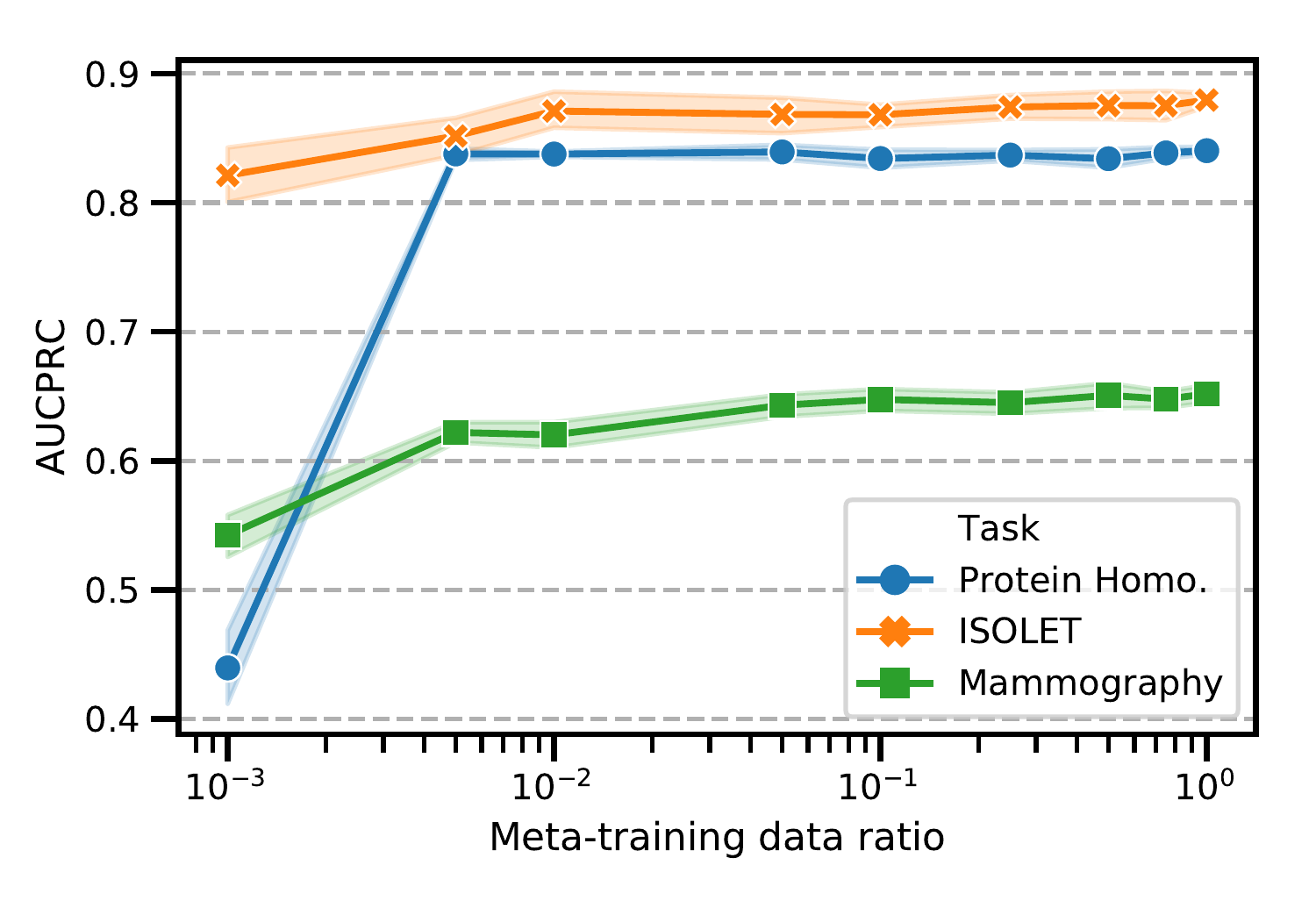}
  }
  \subfigure[Sub-task meta-training time.]{
		\centering
		\includegraphics[width=0.48\linewidth]{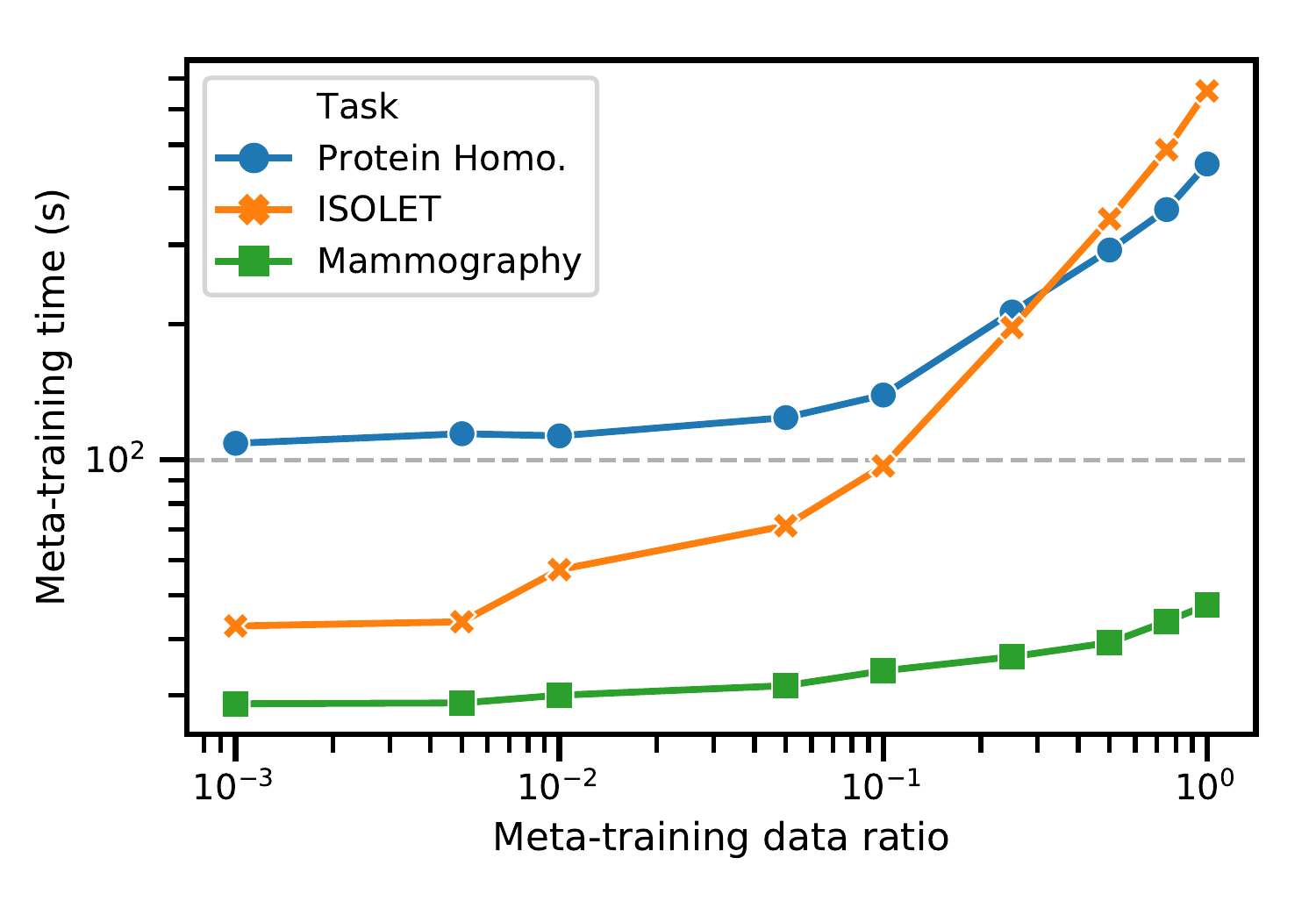}
  }
  \caption{The influence of scaling down the meta-training set.}
  \label{fig:subtask-meta-training-cost}
\end{figure}

Note that, reducing the number of meta-training instances only influences the $C_{f\text{train}}(\cdot)$ term. 
Therefore, the larger the $C_{f\text{train}}(2|\mathcal{P}|)/C_{\Im \text{update}}$, the higher the acceleration ratio brought by shrinking the meta-training set.
We also show some results in Fig.~\ref{fig:subtask-meta-training-cost} to demonstrate such influence. 
The decision tree classifier we used has no max depth limitation, thus its training cost is higher when dealing with high-dimensional data.
We thus choose three tasks with different numbers of features for the test, they are {\it Mammography}/{\it Protein Homo.}/{\it ISOLET} with 6/74/617 features.
It can be observed that the acceleration effect is slightly weaker for the low-dimensional {\it Mammography} task, as the cost of training base classifier is small compared with the cost of updating meta-sampler.
On the other hand, for those high-dimensional tasks (i.e., {\it ISOLET} and {\it Protein Homo.}), shrinking the meta-training set greatly reduces the cost of meta-sampler training as we expect.

\subsection{Guideline of selecting {\sc Mesa} hyper-parameters}
\label{section:hyper-parameter}
The meta-state size $b$ determines how detailed our error distribution approximation is (i.e., the number of bins in the histogram). 
Thus setting a small meta-state size may lead to poor performance. 
Increasing it to a large value (e.g., $\ge$20) brings greater computational cost but only trivial performance increment. 
We recommend setting the meta state size to be 10. 
One can try a bigger meta-state when working on larger datasets. 

% The ensemble size $k$ in meta-training determines the maximum number of base classifiers in an ensemble in each meta-training episode. 
% This number should be set to an appropriate value that enables the ensemble classifier $F_k$ to fit the training set, i.e., adding more base learners into the $F_k$ cannot further improve its generalized performance.

The Gaussian function parameter $\sigma$ determines how to execute meta-sampling in {\sc Mesa}. 
Specifically, given an action $\mu$, we expect the meta-sampling selects those instances with error values close to $\mu$. 
Besides, the meta-sampling is also responsible for providing diversity, which is an important characteristic in classifiers combination. 
A small $\sigma$ can guarantee to select examples with small errors around $\mu$, but this would result in subsets that lack diversity. 
For example, in the late iterations of ensemble training, most of the data instances have stable error values, and meta-sampling with small $\sigma$ will always return the same training set for a specific $\mu$. 
This is detrimental to further improve the ensemble classifier. 
Setting a large $\sigma$ will ``flatten '' the Gaussian function, more instances with different errors are likely to be selected and thus bring more diversity. 
However, when $\sigma \to \infty$, the meta-sampling turns into uniform random under-sampling that makes no sense for meta-training. 
We also note that although one can expand the policy to automatically determine $\sigma$, it requires additional computational cost and the benefit is very limited.
More importantly, selecting inappropriate $\sigma$ will interfere with the quality of collected transitions, causing an unstable meta-training process.
Therefore, we suggest using $\sigma=0.2$ to balance between these factors. 

\section{Visualization}

\begin{figure}[t]
  \centering
  \includegraphics[width=1.0\linewidth]{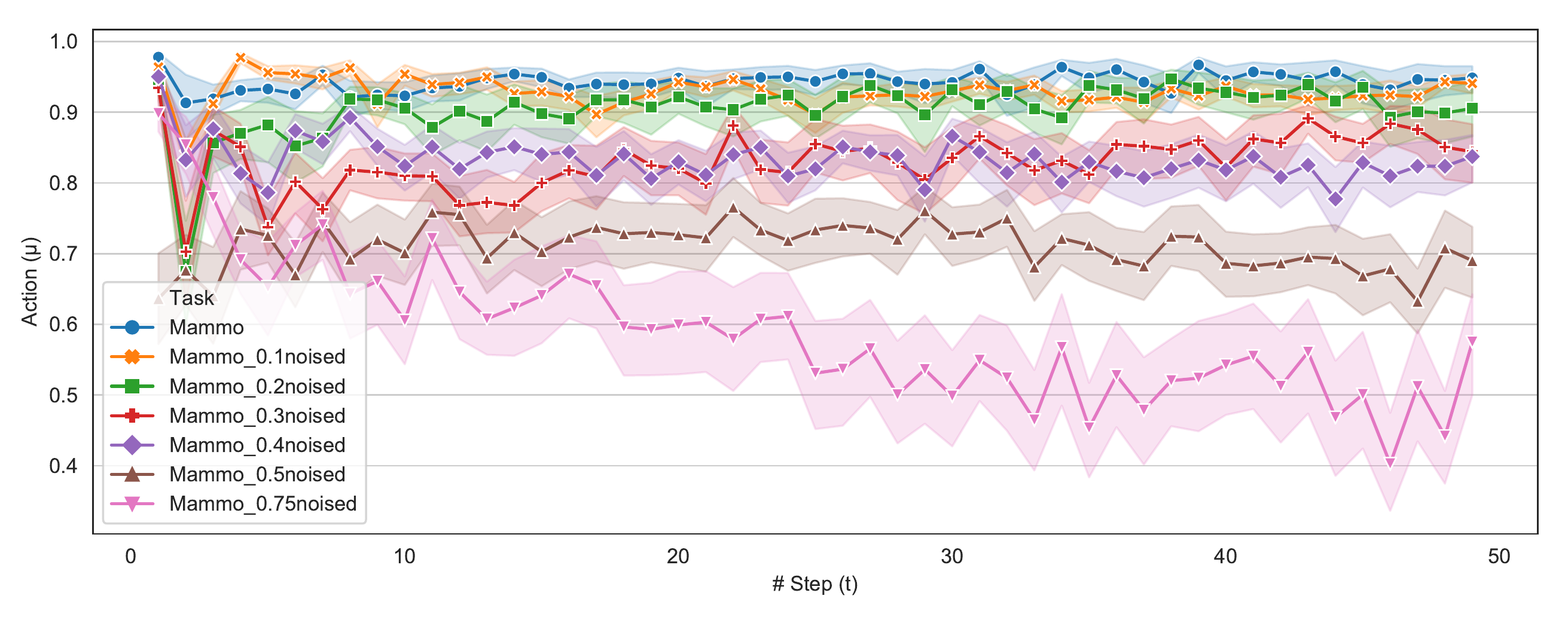}
  \caption{Learned meta-sampler policies on {\it Mammography} dataset with varying label noise ratios.}
  \label{fig:action-under-noises}
\end{figure}

\subsection{Visualization of learned meta-sampler policy}
We visualize the learned meta-sampler policy under different levels of noises in Fig.~\ref{fig:action-under-noises}. 
It clearly shows that the sampling strategy becomes more conservative as the noise ratio grows. 
At the very start of ensemble training, there are only a few base learners and thus the ensemble classifier underfits the training set. 
At this point, the meta-sampler tends to select training instances with larger errors, hence accelerating the fitting process. 
It continues to use such a strategy on datasets with no/few noises. 
However, on highly noisy datasets (e.g., with label noise ratio $\ge$ 40\%), the meta-sampler prefers to select training instances with relatively lower errors in later iterations as the hard-to-classify instances are likely to be noises/outliers. 
This effectively prevents the ensemble classifier from overfitting noisy data points. 

\subsection{Visualization of meta-training process}
\begin{figure}[t]
  \centering
  \includegraphics[width=1.0\linewidth]{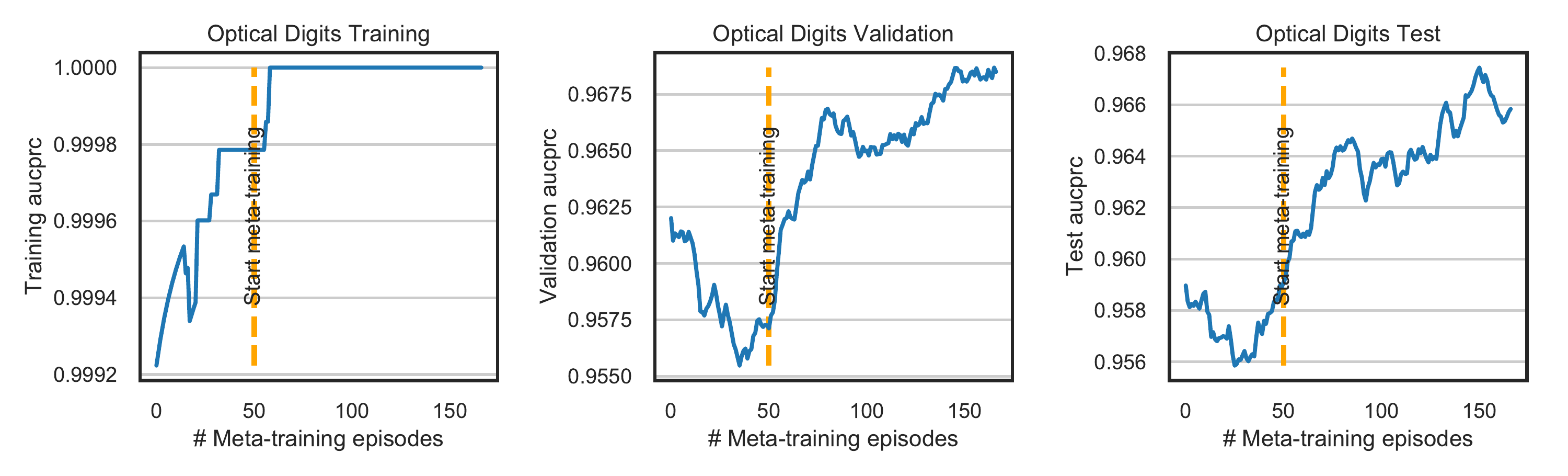}
  \includegraphics[width=1.0\linewidth]{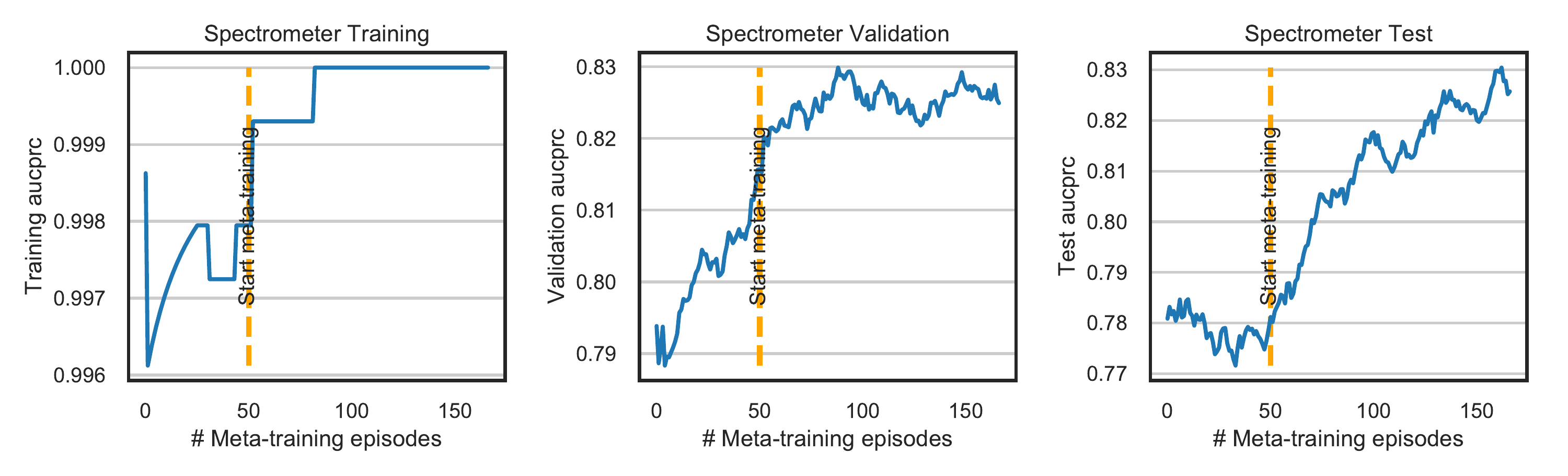}
  \includegraphics[width=1.0\linewidth]{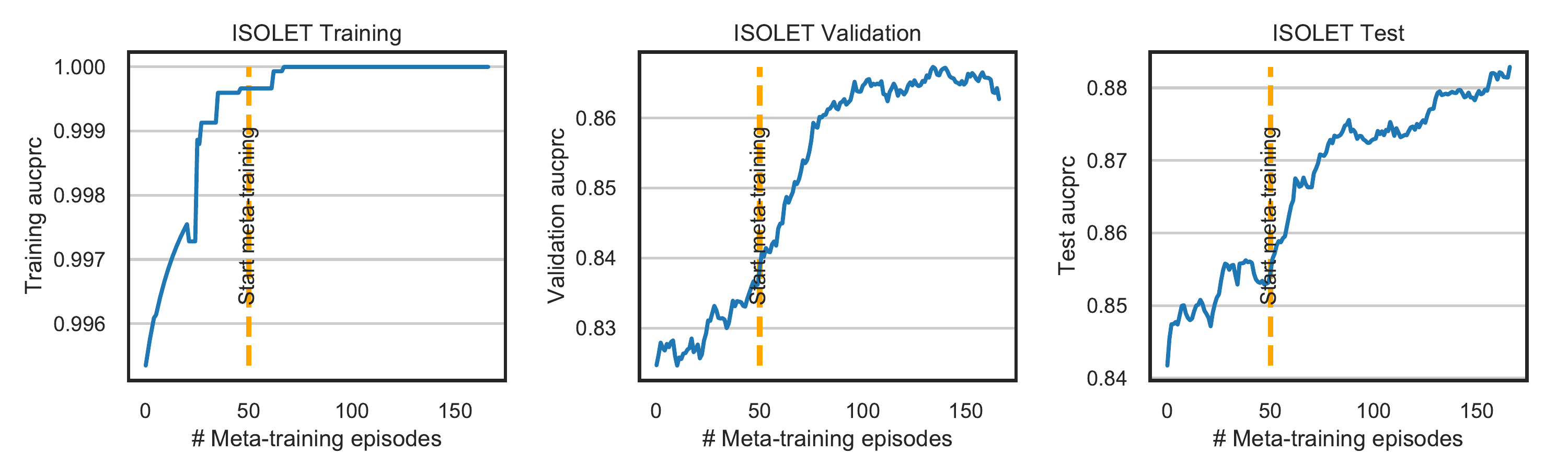}
  \includegraphics[width=1.0\linewidth]{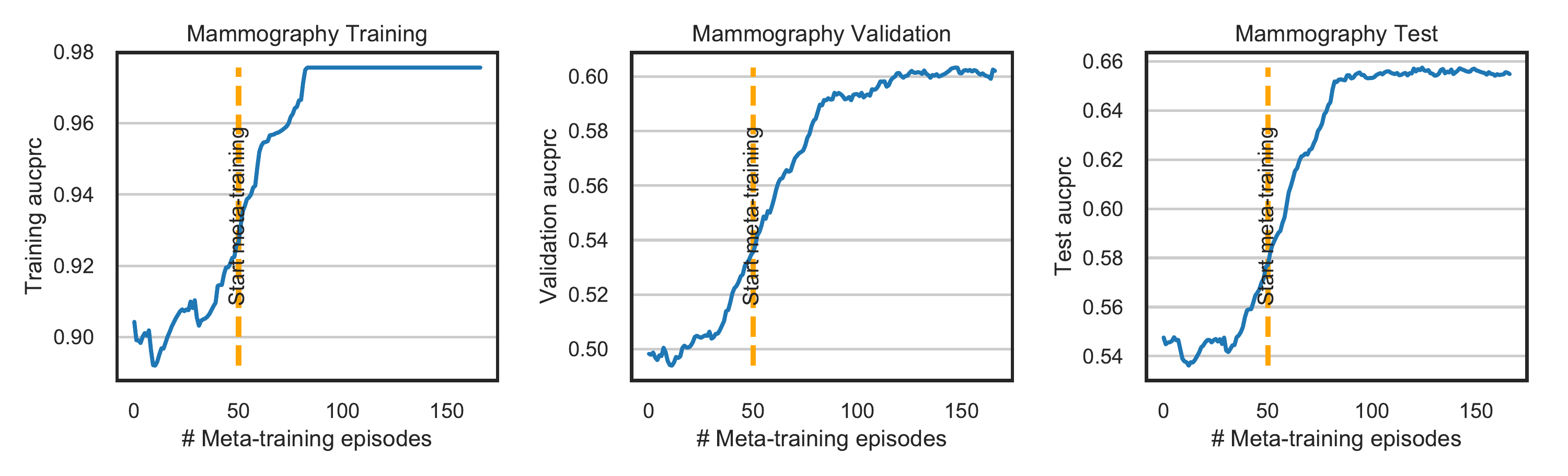}
  \includegraphics[width=1.0\linewidth]{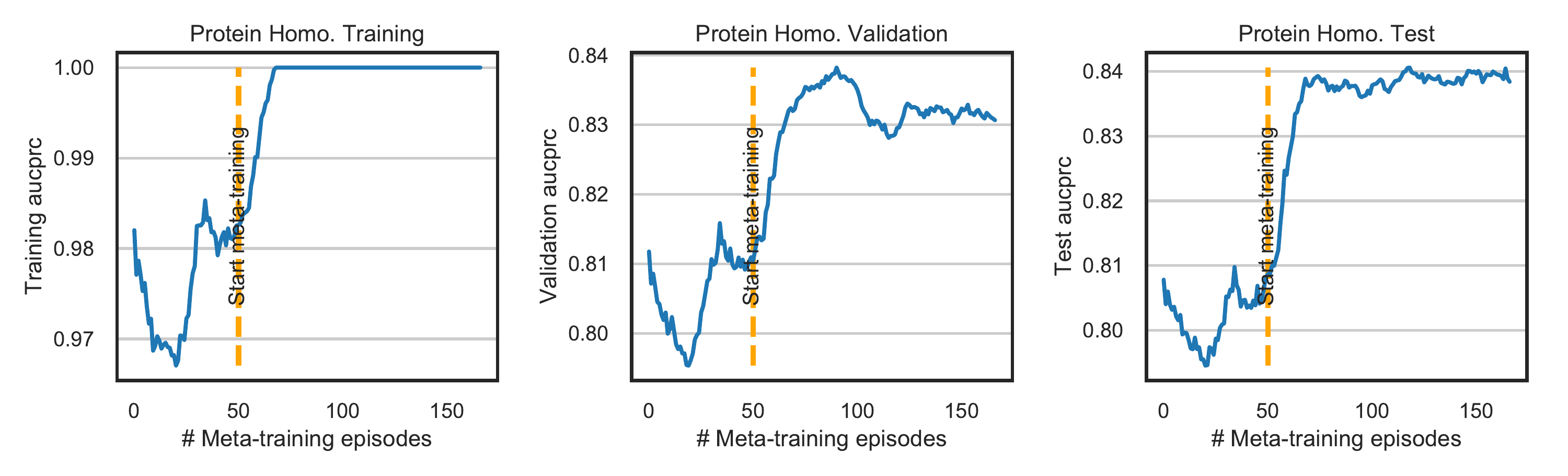}
  \caption{Train/Validation/Test performance during meta-training process (slide mean window=50).}
  \label{fig:visualization-meta-training-process}
\end{figure}

We visualize the meta-training process in Fig.~\ref{fig:visualization-meta-training-process}. 
As the meta-training progress, the classification performance shows consistent improvement in training, validation, and test set in all tasks. 

%% file: mesa.bbl
\begin{thebibliography}{10}

\bibitem{albert02018experiment}
Fernández Alberto, García Salvador, Galar Mikel, Prati Ronaldo~C., and
  Krawczyk Bartosz.
\newblock {\em Learning from Imbalanced Data Sets}.
\newblock Springer, 2018.

\bibitem{barandela2003underbagging}
Ricardo Barandela, Rosa~Maria Valdovinos, and Jos{\'e}~Salvador S{\'a}nchez.
\newblock New applications of ensembles of classifiers.
\newblock {\em Pattern Analysis \& Applications}, 6(3):245--256, 2003.

\bibitem{batista2003smotetomek}
Gustavo~EAPA Batista, Ana~LC Bazzan, and Maria~Carolina Monard.
\newblock Balancing training data for automated annotation of keywords: a case
  study.
\newblock In {\em WOB}, pages 10--18, 2003.

\bibitem{batista2004smoteenn}
Gustavo~EAPA Batista, Ronaldo~C Prati, and Maria~Carolina Monard.
\newblock A study of the behavior of several methods for balancing machine
  learning training data.
\newblock {\em ACM SIGKDD explorations newsletter}, 6(1):20--29, 2004.

\bibitem{chai2004csnb}
Xiaoyong Chai, Lin Deng, Qiang Yang, and Charles~X Ling.
\newblock Test-cost sensitive naive bayes classification.
\newblock In {\em Fourth IEEE International Conference on Data Mining
  (ICDM'04)}, pages 51--58. IEEE, 2004.

\bibitem{chawla2002smote}
Nitesh~V Chawla, Kevin~W Bowyer, Lawrence~O Hall, and W~Philip Kegelmeyer.
\newblock Smote: synthetic minority over-sampling technique.
\newblock {\em Journal of artificial intelligence research}, 16:321--357, 2002.

\bibitem{chawla2003smoteboost}
Nitesh~V Chawla, Aleksandar Lazarevic, Lawrence~O Hall, and Kevin~W Bowyer.
\newblock Smoteboost: Improving prediction of the minority class in boosting.
\newblock In {\em European conference on principles of data mining and
  knowledge discovery}, pages 107--119. Springer, 2003.

\bibitem{chen2010ramoboost}
Sheng Chen, Haibo He, and Edwardo~A Garcia.
\newblock Ramoboost: ranked minority oversampling in boosting.
\newblock {\em IEEE Transactions on Neural Networks}, 21(10):1624--1642, 2010.

\bibitem{davis2006aucprc}
Jesse Davis and Mark Goadrich.
\newblock The relationship between precision-recall and roc curves.
\newblock In {\em Proceedings of the 23rd international conference on Machine
  learning}, pages 233--240. ACM, 2006.

\bibitem{Dua2019uci}
Dheeru Dua and Casey Graff.
\newblock {UCI} machine learning repository, 2017.

\bibitem{finn2017model-agnostic-meta-learning}
Chelsea Finn, Pieter Abbeel, and Sergey Levine.
\newblock Model-agnostic meta-learning for fast adaptation of deep networks.
\newblock In {\em Proceedings of the 34th International Conference on Machine
  Learning-Volume 70}, pages 1126--1135. JMLR. org, 2017.

\bibitem{freund1997adaboost}
Yoav Freund and Robert~E Schapire.
\newblock A decision-theoretic generalization of on-line learning and an
  application to boosting.
\newblock {\em Journal of computer and system sciences}, 55(1):119--139, 1997.

\bibitem{graepel2010ctr}
Thore Graepel, Joaquin~Quinonero Candela, Thomas Borchert, and Ralf Herbrich.
\newblock Web-scale bayesian click-through rate prediction for sponsored search
  advertising in microsoft's bing search engine.
\newblock Omnipress, 2010.

\bibitem{haarnoja2018soft-actor-critic}
Tuomas Haarnoja, Aurick Zhou, Pieter Abbeel, and Sergey Levine.
\newblock Soft actor-critic: Off-policy maximum entropy deep reinforcement
  learning with a stochastic actor.
\newblock In {\em International Conference on Machine Learning}, pages
  1861--1870, 2018.

\bibitem{haixiang2017learning-from-imb-review}
Guo Haixiang, Li~Yijing, Jennifer Shang, Gu~Mingyun, Huang Yuanyue, and Gong
  Bing.
\newblock Learning from class-imbalanced data: Review of methods and
  applications.
\newblock {\em Expert Systems with Applications}, 73:220--239, 2017.

\bibitem{han2005borderline-smote}
Hui Han, Wen-Yuan Wang, and Bing-Huan Mao.
\newblock Borderline-smote: a new over-sampling method in imbalanced data sets
  learning.
\newblock In {\em International conference on intelligent computing}, pages
  878--887. Springer, 2005.

\bibitem{he2008adasyn}
Haibo He, Yang Bai, Edwardo~A Garcia, and Shutao Li.
\newblock Adasyn: Adaptive synthetic sampling approach for imbalanced learning.
\newblock In {\em 2008 IEEE International Joint Conference on Neural Networks
  (IEEE World Congress on Computational Intelligence)}, pages 1322--1328. IEEE,
  2008.

\bibitem{he2008overview}
Haibo He and Edwardo~A Garcia.
\newblock Learning from imbalanced data.
\newblock {\em IEEE Transactions on Knowledge \& Data Engineering},
  (9):1263--1284, 2008.

\bibitem{he2013overview}
Haibo He and Yunqian Ma.
\newblock {\em Imbalanced learning: foundations, algorithms, and applications}.
\newblock John Wiley \& Sons, 2013.

\bibitem{hendrycks2018using-trusted-data-noisy-labels}
Dan Hendrycks, Mantas Mazeika, Duncan Wilson, and Kevin Gimpel.
\newblock Using trusted data to train deep networks on labels corrupted by
  severe noise.
\newblock In {\em Advances in neural information processing systems}, pages
  10456--10465, 2018.

\bibitem{japkowicz2002systematic-study}
Nathalie Japkowicz and Shaju Stephen.
\newblock The class imbalance problem: A systematic study.
\newblock {\em Intelligent data analysis}, 6(5):429--449, 2002.

\bibitem{jiang2017mentornet}
Lu~Jiang, Zhengyuan Zhou, Thomas Leung, Li-Jia Li, and Li~Fei-Fei.
\newblock Mentornet: Learning data-driven curriculum for very deep neural
  networks on corrupted labels.
\newblock {\em arXiv preprint arXiv:1712.05055}, 2017.

\bibitem{krawczyk2016learning}
Bartosz Krawczyk.
\newblock Learning from imbalanced data: open challenges and future directions.
\newblock {\em Progress in Artificial Intelligence}, 5(4):221--232, 2016.

\bibitem{kubat1997oss}
Miroslav Kubat, Stan Matwin, et~al.
\newblock Addressing the curse of imbalanced training sets: one-sided
  selection.
\newblock In {\em Icml}, volume~97, pages 179--186. Nashville, USA, 1997.

\bibitem{lake2015meta-learning}
Brenden~M Lake, Ruslan Salakhutdinov, and Joshua~B Tenenbaum.
\newblock Human-level concept learning through probabilistic program induction.
\newblock {\em Science}, 350(6266):1332--1338, 2015.

\bibitem{laurikkala2001ncr}
Jorma Laurikkala.
\newblock Improving identification of difficult small classes by balancing
  class distribution.
\newblock In {\em Conference on Artificial Intelligence in Medicine in Europe},
  pages 63--66. Springer, 2001.

\bibitem{guillaume2017imblearn}
Guillaume Lema{{\^i}}tre, Fernando Nogueira, and Christos~K. Aridas.
\newblock Imbalanced-learn: A python toolbox to tackle the curse of imbalanced
  datasets in machine learning.
\newblock {\em Journal of Machine Learning Research}, 18(17):1--5, 2017.

\bibitem{li2019gradient-harmonize}
Buyu Li, Yu~Liu, and Xiaogang Wang.
\newblock Gradient harmonized single-stage detector.
\newblock In {\em Proceedings of the AAAI Conference on Artificial
  Intelligence}, volume~33, pages 8577--8584, 2019.

\bibitem{li2017learning-from-noisy-labels}
Yuncheng Li, Jianchao Yang, Yale Song, Liangliang Cao, Jiebo Luo, and Li-Jia
  Li.
\newblock Learning from noisy labels with distillation.
\newblock In {\em Proceedings of the IEEE International Conference on Computer
  Vision}, pages 1910--1918, 2017.

\bibitem{lin2017focalloss}
Tsung-Yi Lin, Priya Goyal, Ross Girshick, Kaiming He, and Piotr Doll{\'a}r.
\newblock Focal loss for dense object detection.
\newblock In {\em Proceedings of the IEEE international conference on computer
  vision}, pages 2980--2988, 2017.

\bibitem{ling2004csdt}
Charles~X Ling, Qiang Yang, Jianning Wang, and Shichao Zhang.
\newblock Decision trees with minimal costs.
\newblock In {\em Proceedings of the twenty-first international conference on
  Machine learning}, page~69. ACM, 2004.

\bibitem{liu2009ee-bc}
Xu-Ying Liu, Jianxin Wu, and Zhi-Hua Zhou.
\newblock Exploratory undersampling for class-imbalance learning.
\newblock {\em IEEE Transactions on Systems, Man, and Cybernetics, Part B
  (Cybernetics)}, 39(2):539--550, 2009.

\bibitem{liu2006cost-sensitive-imbalance}
Xu-Ying Liu and Zhi-Hua Zhou.
\newblock The influence of class imbalance on cost-sensitive learning: An
  empirical study.
\newblock In {\em Sixth International Conference on Data Mining (ICDM'06)},
  pages 970--974. IEEE, 2006.

\bibitem{liu2019self-paced-ensemble}
Zhining Liu, Wei Cao, Zhifeng Gao, Jiang Bian, Hechang Chen, Yi~Chang, and
  Tie-Yan Liu.
\newblock Self-paced ensemble for highly imbalanced massive data
  classification.
\newblock In {\em 2020 IEEE 36th International Conference on Data Engineering
  (ICDE)}. IEEE, 2020.

\bibitem{mani2003nearmiss}
Inderjeet Mani and I~Zhang.
\newblock knn approach to unbalanced data distributions: a case study involving
  information extraction.
\newblock In {\em Proceedings of workshop on learning from imbalanced
  datasets}, volume 126, 2003.

\bibitem{pedregosa2011sklearn}
Fabian Pedregosa, Ga{\"e}l Varoquaux, Alexandre Gramfort, Vincent Michel,
  Bertrand Thirion, Olivier Grisel, Mathieu Blondel, Peter Prettenhofer, Ron
  Weiss, Vincent Dubourg, et~al.
\newblock Scikit-learn: Machine learning in python.
\newblock {\em Journal of machine learning research}, 12(Oct):2825--2830, 2011.

\bibitem{peng2019trainable-under-sampling}
Minlong Peng, Qi~Zhang, Xiaoyu Xing, Tao Gui, Xuanjing Huang, Yu-Gang Jiang,
  Keyu Ding, and Zhigang Chen.
\newblock Trainable undersampling for class-imbalance learning.
\newblock 2019.

\bibitem{ren2018learning-to-reweight}
Mengye Ren, Wenyuan Zeng, Bin Yang, and Raquel Urtasun.
\newblock Learning to reweight examples for robust deep learning.
\newblock In {\em International Conference on Machine Learning}, pages
  4334--4343, 2018.

\bibitem{seiffert2010rusboost}
Chris Seiffert, Taghi~M Khoshgoftaar, Jason Van~Hulse, and Amri Napolitano.
\newblock Rusboost: A hybrid approach to alleviating class imbalance.
\newblock {\em IEEE Transactions on Systems, Man, and Cybernetics-Part A:
  Systems and Humans}, 40(1):185--197, 2010.

\bibitem{shrivastava2016hard-example-mining}
Abhinav Shrivastava, Abhinav Gupta, and Ross Girshick.
\newblock Training region-based object detectors with online hard example
  mining.
\newblock In {\em Proceedings of the IEEE conference on computer vision and
  pattern recognition}, pages 761--769, 2016.

\bibitem{han2018meta-weight-net}
Jun Shu, Qi~Xie, Lixuan Yi, Qian Zhao, Sanping Zhou, Zongben Xu, and Deyu Meng.
\newblock Meta-weight-net: Learning an explicit mapping for sample weighting.
\newblock In {\em NeurIPS}, 2019.

\bibitem{smith2014instance-complexity}
Michael~R Smith, Tony Martinez, and Christophe Giraud-Carrier.
\newblock An instance level analysis of data complexity.
\newblock {\em Machine learning}, 95(2):225--256, 2014.

\bibitem{sun2007cost-boost}
Yanmin Sun, Mohamed~S Kamel, Andrew~KC Wong, and Yang Wang.
\newblock Cost-sensitive boosting for classification of imbalanced data.
\newblock {\em Pattern Recognition}, 40(12):3358--3378, 2007.

\bibitem{tomek1976allknn}
Ivan Tomek.
\newblock An experiment with the edited nearest-neighbor rule.
\newblock {\em IEEE Transactions on systems, Man, and Cybernetics},
  (6):448--452, 1976.

\bibitem{tomek1976tomeklink}
Ivan Tomek.
\newblock Two modifications of cnn.
\newblock {\em IEEE Trans. Systems, Man and Cybernetics}, 6:769--772, 1976.

\bibitem{wang2009smotebagging}
Shuo Wang and Xin Yao.
\newblock Diversity analysis on imbalanced data sets by using ensemble models.
\newblock In {\em 2009 IEEE Symposium on Computational Intelligence and Data
  Mining}, pages 324--331. IEEE, 2009.

\bibitem{wilson1972enn}
Dennis~L Wilson.
\newblock Asymptotic properties of nearest neighbor rules using edited data.
\newblock {\em IEEE Transactions on Systems, Man, and Cybernetics},
  (3):408--421, 1972.

\bibitem{wu2018learning2teach}
Lijun Wu, Fei Tian, Yingce Xia, Yang Fan, Tao Qin, Lai Jian-Huang, and Tie-Yan
  Liu.
\newblock Learning to teach with dynamic loss functions.
\newblock In {\em Advances in Neural Information Processing Systems}, pages
  6466--6477, 2018.

\end{thebibliography}
